\documentclass[10pt,twocolumn,letterpaper]{article}

\usepackage[pagenumbers]{cvpr} 

\usepackage[dvipsnames]{xcolor}
\usepackage{verbatim}

\usepackage{makecell}
\usepackage{graphicx}
\usepackage{multirow}
\usepackage{float}
\usepackage{caption}
\usepackage{multicol}
\usepackage{placeins}
\usepackage{color, colortbl}
\definecolor{Gray}{gray}{0.9}
\definecolor{Yellow}{rgb}{1.0, 0.75, 0.0}\usepackage{amssymb} 
\usepackage{pifont}  
\newcommand{\cmark}{\ding{51}}
\newcommand{\xmark}{\ding{55}}
\usepackage{arydshln}
\usepackage{booktabs}
\usepackage{makecell}
\usepackage{graphicx}
\usepackage{multirow}
\usepackage{float}
\usepackage{caption}
\usepackage{multicol}
\usepackage{placeins}
\usepackage{tabu}
\usepackage{color, colortbl}
\usepackage[symbol]{footmisc}
\usepackage{soul}

\definecolor{Gray}{gray}{0.9}
\definecolor{dGray}{gray}{0.6}
\definecolor{OliveGreen}{rgb}{0,0.6,0}
\definecolor{red}{rgb}{0.6,0,0}

\usepackage{xcolor}

\usepackage[normalem]{ulem}

\newcommand{\methodname}{TAP-VL}
\newcommand{\methodnamelw}{$\text{TAP-VL}_{\text{Light}}$}
\newcommand{\ptname}{Text Layout-Aware Pretraining}
\newcommand{\qfname}{OCR-Q}
\newcommand{\pttwo}{OCR-Mask Contrastive Learning}
\newcommand{\ptthree}{OCR-Mask Matching}
\newcommand{\ptone}{OCR-Grounded Mask Denoising}

\usepackage{amsmath,amsfonts,bm}

\def\eqref#1{equation~\ref{#1}}

\def\1{\bm{1}}

\def\rvb{{\mathbf{b}}}

\def\rmP{{\mathbf{P}}}

\def\rmR{{\mathbf{R}}}
\def\rmS{{\mathbf{S}}}

\def\evt{{t}}

\DeclareMathAlphabet{\mathsfit}{\encodingdefault}{\sfdefault}{m}{sl}
\SetMathAlphabet{\mathsfit}{bold}{\encodingdefault}{\sfdefault}{bx}{n}

\newcommand{\R}{\mathbb{R}}

\definecolor{cvprblue}{rgb}{0.21,0.49,0.74}
\usepackage[pagebackref,breaklinks,colorlinks,citecolor=cvprblue]{hyperref}

\def\paperID{*****} 
\def\confName{CVPR}
\def\confYear{2025}

\title{TAP-VL: Text Layout Aware Pretraining for Enriched Vision-Language Models}

\author{
Jonathan Fhima\\
Technion, Israel\\
\and
Elad Ben Avraham\\
AWS AI Labs\\
\and
Oren Nuriel\\
AWS AI Labs\\
\and
Yair Kittenplon\\
AWS AI Labs\\
\and
Roy Ganz\\
AWS AI Labs\\
\and
Aviad Aberdam\\
AWS AI Labs\\
\and
Ron Litman\\
AWS AI Labs\\
}

\begin{document}
\maketitle

\begin{abstract}
Vision-Language (VL) models have garnered considerable research interest; however, they still face challenges in effectively handling text within images. To address this limitation, researchers have  developed two approaches. The first method involves utilizing external Optical Character Recognition (OCR) tools to extract textual information from images and prepend it to the textual inputs. The second strategy is OCR-free and focuses on employing extremely high-resolution images to improve text recognition capabilities. In this paper, we focus on enhancing the first strategy by introducing a novel method, named \methodname{}, which treats OCR information as a distinct modality and seamlessly integrates it into any VL model. \methodname{} employs a lightweight transformer-based OCR module to receive OCR with layout information, compressing it into a short fixed-length sequence which serves as an input for the LLM. To this end, we conduct model-agnostic pretraining of the OCR module on unlabeled documents, followed by its integration into any VL architecture through short fine-tuning. Extensive experiments demonstrate consistent performance improvements when applying \methodname{} to top-performing VL models, across scene-text and document-based benchmarks.

\end{abstract}
\section{Introduction}
\label{sec:intro}

Large Vision-Language (VL) models have emerged as a key research area in the field of artificial intelligence, leading to significant progress in multimodal reasoning \cite{alayrac2022flamingo,dai2023instructblip,ganz2024clipag,ganz2023models,hu2023bliva,li2022blip,Li2023BLIP2BL,liu2023llava,bai2023qwenvl,chen2023pali,ganz2024question,litmanvisfocus, blau2024gram}. Such architectures bridge the gap between visual and textual data by integrating a vision encoder and a Large Language Model (LLM) via a translation module. This module projects the visual encodings into the text embeddings space. As VL models can generate content based on both visual and textual information, they play a pivotal role in a diverse set of applications and tasks, including image captioning (CAPS)~\cite{chen2015microsoft} and visual question answering (VQA)~\cite{antol2015vqa}.
While open-source VL models have shown impressive performance across various tasks, many still face challenges when dealing with with OCR-oriented tasks such as TextVQA \cite{textvqa}, TextCaps \cite{textcaps}, and DocVQA \cite{docvqa}.
There are two main strategies to address this challenge: (1) integrating an external OCR system to extract OCR tokens and use them as additional input, and (2) employing very high-resolution images combined with extensive pre-training to enhance text recognition. Each approach has its own advantages and limitations, and both are active areas of research. In this paper, we focus on the first approach.

\begin{figure}[t] 
  \centering
  \includegraphics[width=0.9\linewidth]{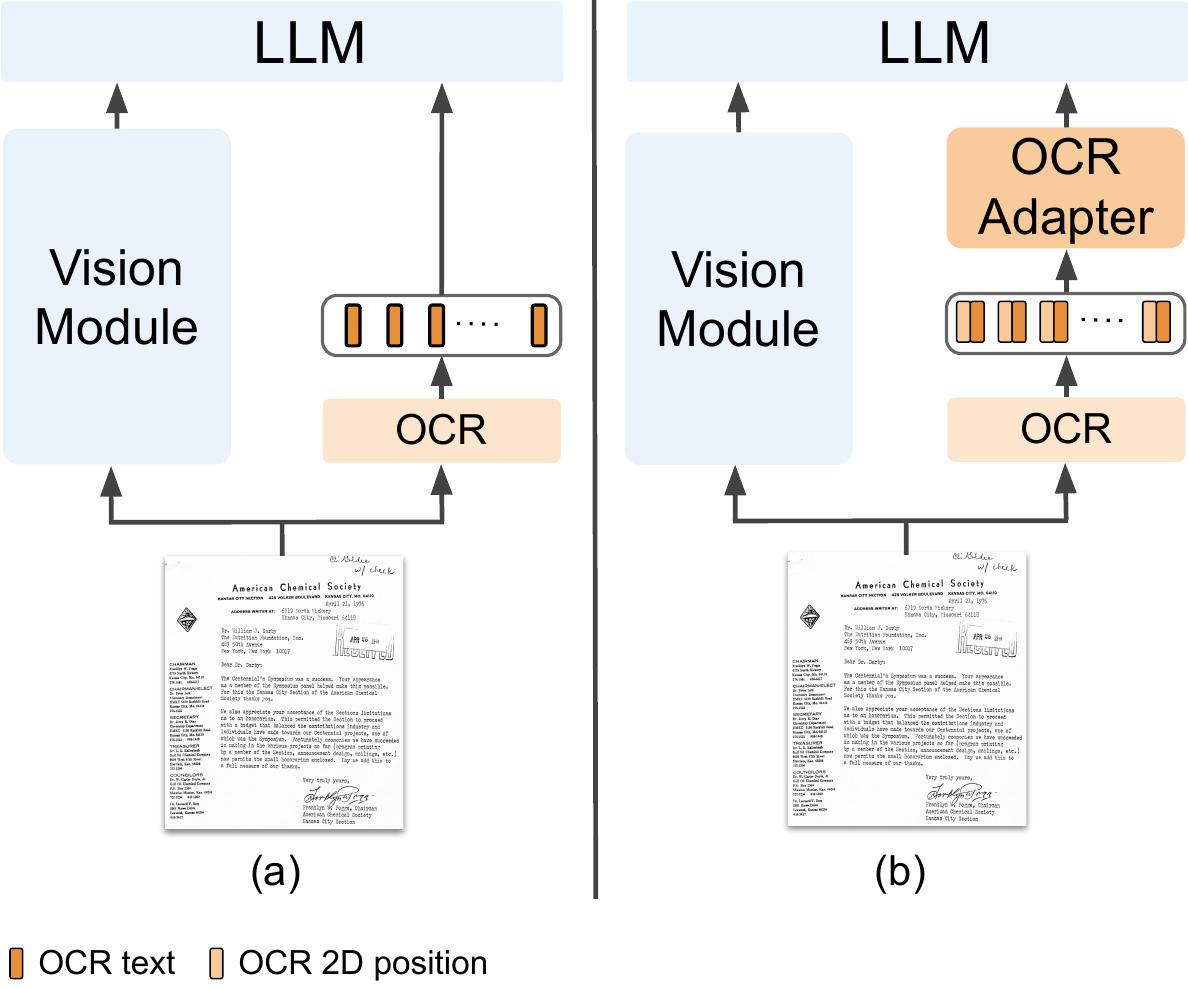}
  \vspace{-0.2cm}
  \caption{\textbf{Layout-aware OCR Adapter}. 
  (a) Previous methods extract OCR data and input it into the LLM as plain text. (b) \methodname{} introduces a plug-and-play OCR adapter that leverages layout information and can be seamlessly integrated with any vision-language LLM (VLLM).
  }
    \label{fig:teaser}
  \vspace{-0.3cm}

\end{figure}

The prevailing paradigm for incorporating OCR into VL systems involves prepending raw OCR-extracted words into the LLM  (left side of \cref{fig:teaser}). While this strategy enhances performance on OCR-oriented benchmarks, it exhibits critical shortcomings. Firstly, it relies solely on OCR tokens, neglecting crucial spatial layout information proven highly beneficial in OCR-oriented tasks~\cite{appalaraju2023docformerv2,latr,ganz2023models,hsu2024m3t}. 
Moreover, when applied to domains with text-rich images, inserting lengthy OCR sequences into the LLM results in significant computational overhead due to the quadratic complexity of the attention mechanism. 

\begin{figure*}[ht] 
  \vspace{-0.6cm}
  \centering
  \includegraphics[width=0.8\linewidth]{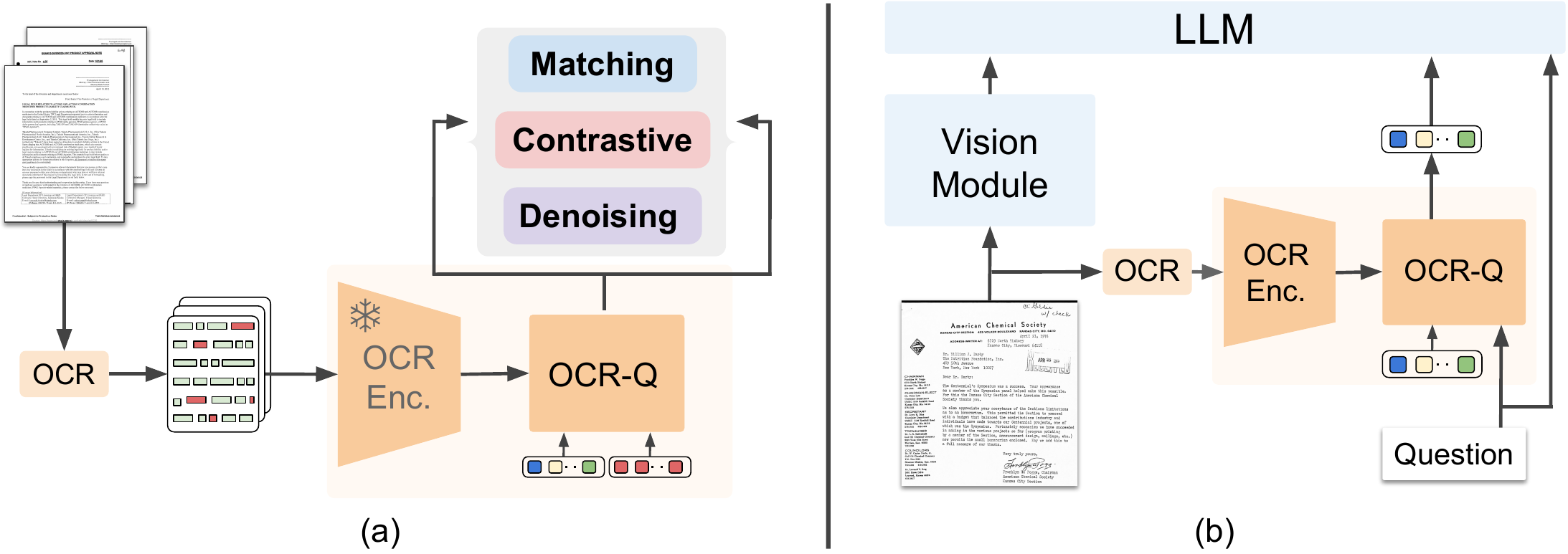}
  \vspace{-0.2cm}
  \caption{\textbf{\methodname{} Approach}. 
  (a) Our model-agnostic layout-aware pretraining framework for creating condensed rich OCR embeddings conditioned on text. (b) \methodname{} fully integrated, enhancing any VL model on OCR-oriented tasks.
  }
    \label{fig:overallarch}
  \vspace{-0.3cm}
\end{figure*}

In this study, we address these limitations and introduce \methodname{}, a technique for seamlessly integrating OCR information into any VL model through short fine-tuning (right side of \cref{fig:teaser}). 
By incorporating 2D positional data in addition to the OCR-extracted word tokens, the model can interpret relationships between different textual elements and understand hierarchical structures that are essential for accurate information extraction and holistic document understanding.
Conceptually, our method treats OCR as a distinct modality and thus employs an OCR module, similar to the use of a dedicated vision module for encoding visual input.
Following this, we introduce a transformer-based lightweight \qfname{}, to generate meaningful representations conditioned on user queries. The OCR encoder captures vital spatial layout information, while the \qfname{} condenses lengthy OCR details into a fixed-size sequence length representation. This condensed representation serves as input for the LLM alongside visual and textual data (right side of \cref{fig:overallarch}). \methodname{} employs these condensed representations to integrate OCR with spatial information into the VL model. 
    
Initially, we introduce a standalone, model-agnostic layout-aware pretraining, as depicted on the left side of \cref{fig:overallarch}. This phase operates independently of the VL model, enhancing efficiency and enabling a focused exploration of OCR understanding without introducing a distribution shift to the VL model.
Aimed at distilling and extracting the most relevant OCR information, we propose a designated layout-aware pretraining that leverages the abundant unlabeled document data with rich layouts and text~\cite{latr,IDL}.
Specifically, we pretrain the \qfname{} in a three-objectives scheme, drawing inspiration from previous works~\cite{latr,appalaraju2023docformerv2,Li2023BLIP2BL}. 
In more detail, our approach consists of the following layout-aware tasks: (1) \ptone{},  which predicts masked spans based on the noisy OCR input; (2) \pttwo{}, which aims to align OCR and word representations within the same document while distinguishing between representations from different documents, and (3) \ptthree{},which aligns noisy OCR text with missing spans. 
Combining such objectives propels the model to acquire a deep layout and OCR understanding while providing a compact representation.

Following this, we integrate the same pretrained model into various leading VL models via a short multi-task fine-tuning procedure. Specifically, we examine prominent VL models such as InstructBLIP \cite{dai2023instructblip}, LLaVA \cite{liu2023llava} and Qwen-VL \cite{bai2023qwenvl}. Our extensive experimentation demonstrates the efficacy of \methodname{} across document understanding and scene-text VL benchmarks, resulting in substantial enhancements compared to diverse baseline methods across all assessed benchmarks, including a zero-shot scenario.


In addition, we propose \methodnamelw{}, a light-weight version of \methodname{} that solely utilizes our compressed OCR representations without providing the LLM with the raw OCR tokens.
This approach is specifically efficient in tasks related to document understanding with dense-text images. Notably, we demonstrate that applying \methodnamelw{} not only significantly reduces the computational costs compared to the relevant baselines (reduces the FLOPs by up to a factor of seven), but also leads to substantial performance improvements. Notably, we showcase \methodnamelw{}'s ability to extrapolate to multi-page scenarios without any specific multi-page training. In the most challenging case of multi-page document understanding, \methodnamelw{} achieves performance improvements of up to 4.8\%, while substantially reducing the computational costs.

In summary, our contributions include:
\begin{itemize}
    \item Introducing \methodname{}, a novel approach for seamlessly integrating OCR information into any pretrained VL model, enabling effective reasoning over both textual and spatial information.
    \item Proposing a unique layout-aware model-agnostic pretraining strategy, utilizing unlabeled document data to acquire rich, condensed OCR features.
    \item Demonstrating the effectiveness of our method in enhancing performance across various state-of-the-art VL architectures, showcasing its ability to elevate performance across multiple benchmarks in scene-text and document understanding tasks, including challenging zero-shot multi-page setting.
    \item We present \methodnamelw{}, a lightweight version of \methodname{}, capable of handling multi-page documents without any specific training. \methodnamelw{} decreases FLOPs by up to four times while still achieving superior performance compared to approaches relying on the uncompressed OCR sequence.
\end{itemize}

\section{Related work}
\subsection{Vision-Language Models}
VL models have undergone significant evolution, transitioning from task-specific to more generalized approaches, facilitated by LLMs \cite{alayrac2022flamingo,panagopoulou2023xinstructblip,chen2023pali,chen2023pali3,chen2023palix,ganz2024question}. These modern models demonstrate adaptability across diverse tasks and impressive generalization capabilities \cite{Li2023BLIP2BL,dai2023instructblip,liu2023llava,chen2023pali}. Architecturally, these models typically involve three fundamental parts. First, a vision architecture extracts meaningful information from images, often employing a frozen vision-transformer as the vision encoder. Second, a translation module bridges the gap between vision and language, transforming visual features into a representation comprehensible and processable by a language model. This module may consist of a simple linear layer or MLP \cite{liu2023llava}, or a cross-attention-based transformer architecture \cite{dai2023instructblip,Li2023BLIP2BL,bai2023qwen}. Lastly, the projected visual information and textual instructions, typically in the form of questions or prompts, are input into an LLM to execute the task. More specifically, BLIP-2\cite{Li2023BLIP2BL} suggest incorporating a Querying-Transformer (Q-Former) to efficiently add visual cues from a vision encoder to an LLM and InstructBLIP \cite{dai2023instructblip} adapts it to follow instructions. LLaVA \cite{liu2023llava, liu2024llavanext} introduce a simple projection layer to translate between the vision modality to the language one and utilize GPT-4 generated multi-modal data. Qwen-VL \cite{bai2023qwenvl} propose a single cross-attention layer and perform a full multi-modal fine-tuning phase to align modalities. Additionally, other works have explored extending these methodologies to encompass multiple modalities, such as audio and video \cite{zhang2023videollama,wang2023onepeace,panagopoulou2023xinstructblip,chen2023xllm}.
\subsection{Integrating OCR information into VL Models}
Before the emergence of large vision language models, numerous methods attempted to address OCR-oriented vision challenges. TAP \cite{yang2021tap} introduced a pretraining objective to align different representations better. In \cite{ganz2023models}, OCR information was integrated into the decoder of an encoder-decoder framework through auxiliary losses. LaTr \cite{latr} proposed a layout-aware transformer, enabling reasoning over textual and layout cues using an unsupervised pretraining. However, these works do not consider recent advances in VL models \cite{dai2023instructblip,bai2023qwenvl} and cannot fully leverage their capabilities.

An alternative approach could target this without assuming an OCR extraction system as the previous approaches do. In this scenario, the VL model would solely rely on visual cues. For instance, OCR-free methods like Qwen-VL and LLaVAR \cite{bai2023qwenvl, zhang2023llavar} observed that open-source all-purpose vision encoders perform poorly in this regard and proposed a training paradigm comprising explicit OCR-oriented tasks. Although performance improved, they still do not surpass methods utilizing OCR systems. Hence, most VL methods \cite{dai2023instructblip,chen2023pali} incorporate these OCR systems.

The conventional method of integrating OCR information into these models involves using the raw text extracted by OCR as part of the input prompt to the LLM \cite{dai2023instructblip,chen2023pali, liu2023llava}. This is feasible as modern VL models can understand that these words are associated with the image, requiring minimal modifications and training to seamlessly integrate OCR-derived text. However, this approach overlooks the fact that OCR extraction also provides word bounding boxes, with layout information being crucial \cite{latr,appalaraju2023docformerv2,ganz2023models}. Additionally, inputting the entire OCR sequence into an LLM is computationally intensive, particularly with OCR-dense images, such as in document understanding tasks, potentially leading to poor performance. Therefore, we introduce \methodname{}, a method that effectively and efficiently incorporates OCR and layout information into any VL model.

\section{Method}
\label{sec:method}



In this section we introduce \methodname{}, which encompasses an OCR module, and a novel layout-aware pretraining paradigm, empowering the VL model with OCR comprehension. The OCR module's architectural design is formulated by treating OCR as an extra, independent modality, addressing its inherent complexity. The layout-aware pretraining aims to teach the OCR module to produce a concise yet rich representation of the OCR. Importantly, this phase operates independently of the VL model, enhancing efficiency and ensuring compatibility with various VL architectures. Following layout-aware pretraining, we integrate our OCR module into any VL architecture through parameter-efficient fine-tuning. This integration leads to significant improvements in OCR benchmarks. Moreover, we propose an additional design choice that can reduce the computational footprint significantly compared to existing approaches, while maintaining comparable performance advantages. In the following sections, we detail our proposed OCR module, the layout-aware model-agnostic pretraining, and the integration into any VL model.

\subsection{Model Architecture}
\label{subsec:architecture}
To enhance the OCR understanding of VL architectures, we propose an OCR module composed of two pivotal components: an OCR encoder and an OCR compressor, which we term \qfname{}. The OCR encoder produces embeddings based on tokens and their 2D positions, encompassing essential layout information. The \qfname{} is a transformer-based module designed to produce a compact representation of the OCR based on the query. It is comprised of two transformer sub-modules which share the same self-attention layers: (i) an OCR transformer module interacting with the encoded OCR embeddings and (ii) a text transformer module, which handles the free-text input (such as a user's question). Specifically, the \qfname{} transforms the OCR embeddings via learnable queries and textual prompt into a fixed number of representations. We define $K$ to be the number of learnable queries used as input to the \qfname{}. Upon integration into a VL model, these compressed representations are concatenated with user instructions and fed into the VL model, as illustrated on the right side of \cref{fig:overallarch}.

\subsection{Layout-Aware pretraining}
The layout-aware pretraining phase aims to produce an \qfname{} capable of compressing OCR content while extracting meaningful information based on textual input. Inspired by the pretraining of BLIP2~\cite{Li2023BLIP2BL}, we adopt a three-objective scheme: \ptone{},  \pttwo{}, and \ptthree{}. To this end, we utilize the myriad of publicly available documents and their OCR information. 
The overall scheme is illustrated on the left side of \cref{fig:overallarch}.

\begin{figure}[tb]  
  \vspace{-0.6cm}
    \centering
   \includegraphics[width=0.95\linewidth]{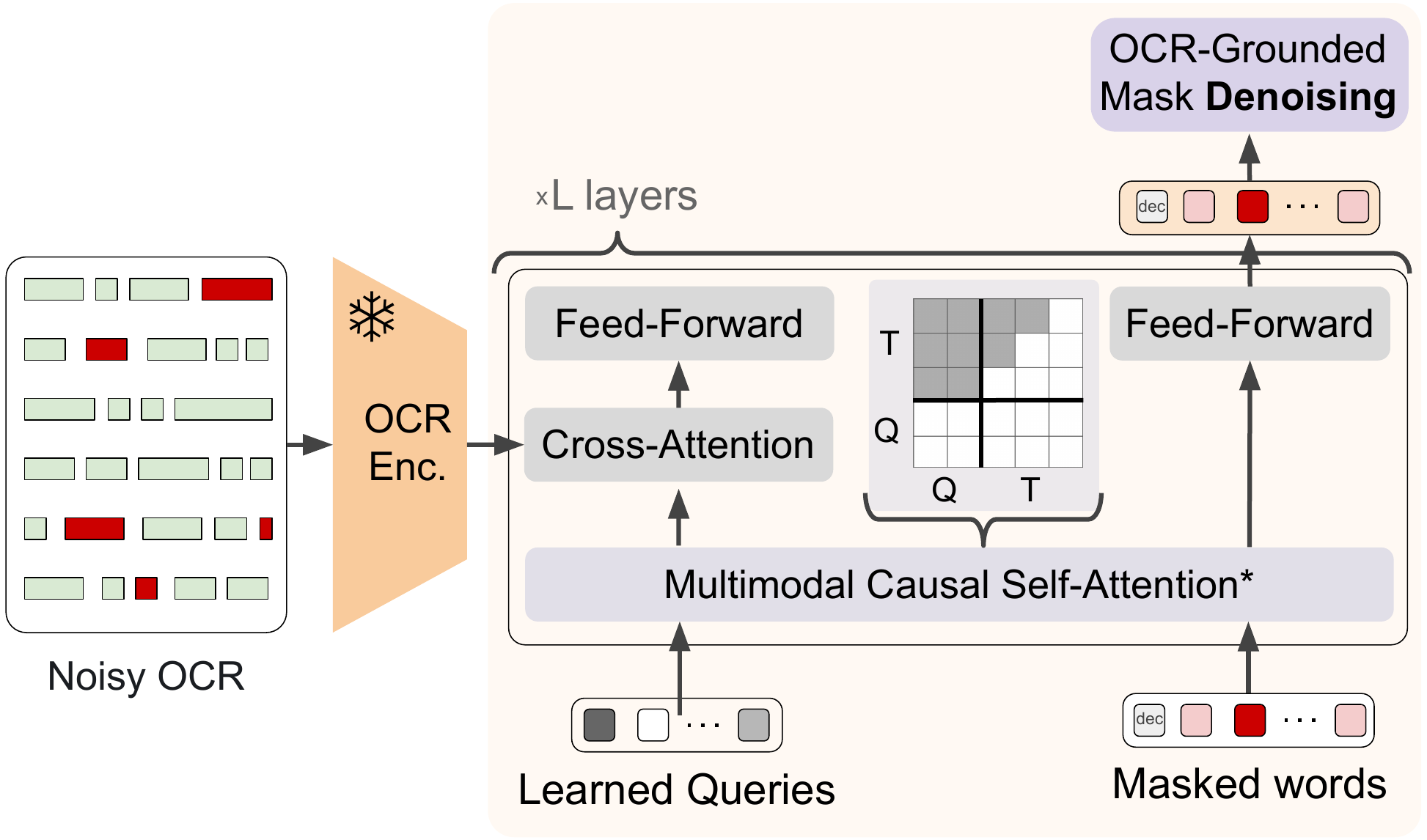}
   \vspace{-0.2cm}

   \caption{\textbf{\ptone{}}. 
  Denoising pretraining mechanism where the \qfname{} predicts the masked words, enhancing comprehension of unmasked text semantics and layout information.
  }
    \label{fig:denoising}
\end{figure}

\begin{figure}[tb]  
  \vspace{-0.6cm}
    \centering
   \includegraphics[width=0.98\linewidth]{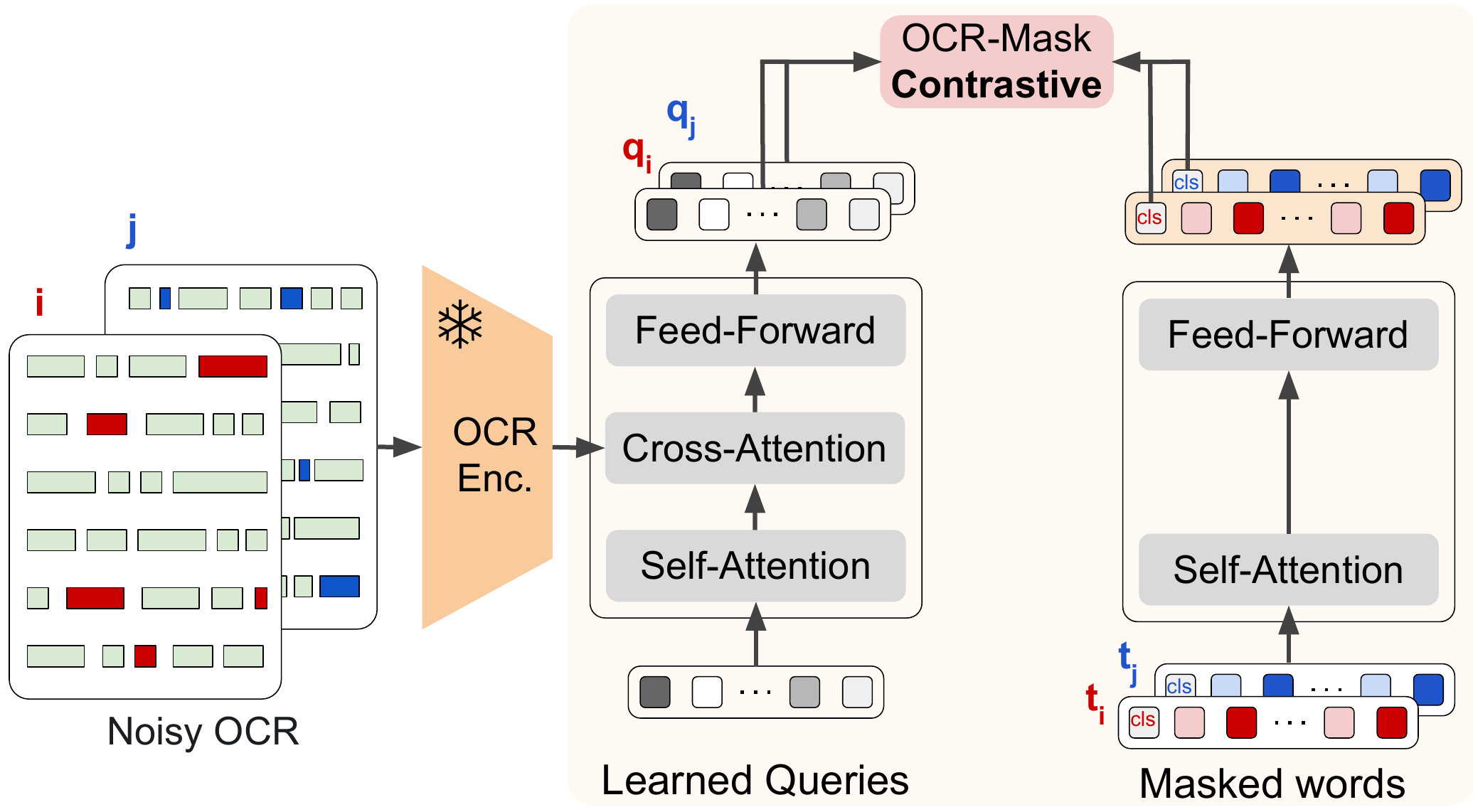}
   \vspace{-0.2cm}

   \caption{\textbf{\pttwo{}}. 
  Contrastive learning task aligning the learnable queries representation (interacting with the masked OCR) with the ones of the masked words.
  }
    \label{fig:contrastive}
\end{figure}

\begin{figure}[t]

    \centering
  \includegraphics[width=0.98\linewidth]{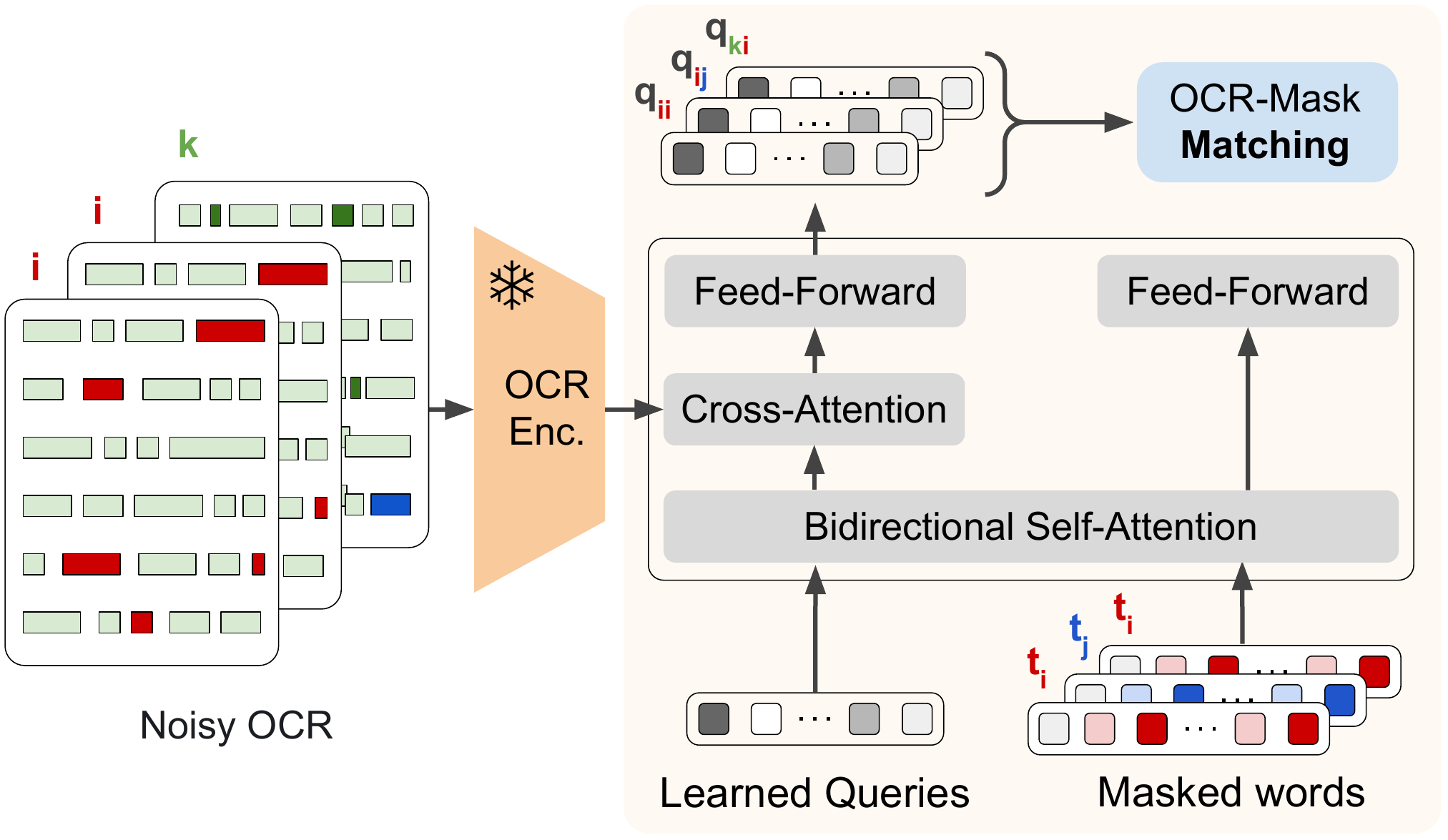}
  
    \vspace{-0.2cm}

  \caption{\textbf{\ptthree{}}. 
  Visualization of the matching phase, aimed at enabling the \qfname{} to determine the correspondence between masked OCR content and the masked words.
  }
    \label{fig:matching}
 \vspace{-0.5cm}
\end{figure}

This approach leverages publicly available unlabeled document corpora by randomly masking spans of OCR, which include both text and layout information. Thus creating pairs consisting of masked versions of the OCR and their corresponding masked words for use during pretraining. 
For each of the pretraining objectives we follow the same general outline (\cref{fig:denoising,fig:contrastive,fig:matching}): the OCR encoder produces rich embeddings given the masked OCR. These embeddings are then fed to the OCR transformer module via the cross-attention layers, resulting in a fixed number of representations per document (determined by $K$). The text transformer module receives the masked words as input and a special token, dependent on the pretraining task. The spatial information of these tokens are omitted as the text transformer module operates on free-form unstructured text. Its output is then fed into a task-specific projection layer, preceding the loss. The interaction between the text and OCR transformer modules is governed by the masking mechanism employed, contingent upon the characteristics of the chosen pretraining objective. Next we elaborate on each pretraining task, for more details please refer to \cref{ssup:pretraining}.

\textbf{\ptone{}}
tasks the \qfname{} with restoring masked words from noisy OCR input, as illustrated in \cref{fig:denoising}. This encourages meaningful compressed representations, leveraging both textual and layout information. 
Throughout this task, the text transformer module indirectly queries the noisy OCR inputs, via the intermediate learned query representations. Since the text transformer module lacks direct access to OCR content, minimizing this loss is feasible only if the representations corresponding to learnable queries are enriched with the relevant OCR information. We use a multimodal mask in our self-attention layers~\cite{dong2019unified,Li2023BLIP2BL} to access compressed OCR information (\cref{fig:denoising}). This mask restricts learnable queries from attending to text tokens but allows interaction among themselves, while text tokens can attend to learnable queries but only self-interact through a causal mask. A special \texttt{<dec>} token is prepended to the masked words as the denoising task prefix.





\textbf{\pttwo{}} aims to align the outputs of the OCR transformer module with the text transformer module of the \qfname{} (\cref{fig:contrastive}). 
The OCR transformer module has access to the noised OCR content, while the text transformer module receives a \texttt{<cls>} special token followed by the masked words. 
The primary purpose of this alignment is to enhance the mutual understanding between OCR-encoded information and textual representations.
In this task, we use a unimodal mask where the learnable queries and the masked words can only attend to themselves.

\textbf{\ptthree{}} task involves a binary classification objective of matching the representations of the compressed noisy OCR information and the ones of the masked words (\cref{fig:matching}). 
Specifically, for a given noisy document, we couple it with its corresponding masked words with probability $p$ and with a non-matching hard negative with probability $1-p$. 
Notably, for this objective, we apply a bidirectional self-attention mask allowing all queries and masked words to attend to each other. 

\subsection{Incorporating OCR QFormer in VL Models}
\label{ssec:finetuning}
Following our model agnostic pretraining, we align our OCR module with any VL model in a two-phase fine-tuning procedure.
First, we employ an \textit{OCR-to-language alignment}, in which we integrate the OCR encoder and \qfname{} to a frozen LLM. In this stage, we train the OCR module to fit the LLM via fine-tuning using OCR-centric VQA datasets.
We use these datasets as most answers can be inferred solely from the OCR information and do not require direct access to the visual inputs~\cite{latr,ganz2023models}.
Next, we conduct an \textit{OCR-vision-to-language alignment} in which we consider the entire VL model, as depicted in \cref{fig:overallarch}.
In this setting, the LLM is fed with textual instructions along with both visual and OCR features, from the OCR and vision modules. In addition to the textual instructions, appending the raw OCR word list, as commonly done in VL works, is an optional design choice. To present the trade-off between the two options we introduce \methodnamelw{}, which omits the raw OCR word list as input to the LLM, making it more computationally efficient than \methodname{} (see \cref{sec:light}).
In this stage, we conduct a multi-task fine-tuning using a mixture of OCR-oriented and non-OCR-oriented VQA and captioning datasets, including both documents and natural images, to align the system's building block.
Specifically, we train the OCR components and employ low-rank adaptation to the LLM~\cite{hu2021lora} while keeping the visual module frozen.
The results in a VL system capable of effectively reasoning over both visual and OCR information.

\begin{table*}[t!]
  \centering
  \resizebox{\linewidth}{!}{%
  \begin{tabular}{c ll|ccc cc c | cc }
    \midrule
    && & \multicolumn{3}{c}{\textbf{Scene-Text}} & \multicolumn{2}{c}{\textbf{Document}} & \textbf{0-shot} & \multicolumn{2}{c}{\textbf{Average}}\\ 
    
    &\textbf{Method} & \textbf{LLM} &  TextVQA \cite{textvqa} & STVQA \cite{stvqa}& TextCaps \cite{textcaps}& DocVQA \cite{docvqa}& InfoVQA \cite{infovqa}& DUDE \cite{dude}& Scene-Text & Document\\
    
    && & \fontsize{7pt}{8pt}\selectfont VQAScore & \fontsize{7pt}{8pt}\selectfont ANLS & \fontsize{7pt}{8pt}\selectfont CIDEr & \fontsize{7pt}{8pt}\selectfont ANLS & \fontsize{7pt}{8pt}\selectfont ANLS & \fontsize{7pt}{8pt}\selectfont ANLS &  &  \\

    \cline{2-11} &  & & & & & & & &\\[-2ex] 
\multirow{6}{*}{\rotatebox{90}{Specialist}} & 
    \selectfont\color{gray}UniTNT \cite{ganz2023models}&\selectfont\color{gray}-& \selectfont\color{gray}55.4 & \selectfont\color{gray}66.0 & \selectfont\color{gray}109.0 & \selectfont\color{gray}- & \selectfont\color{gray}- & \selectfont\color{gray}-&\selectfont\color{gray}- & \selectfont\color{gray}-\\

    &\selectfont\color{gray}DocFormer v2 \cite{appalaraju2023docformerv2}& \selectfont\color{gray}T5 large \cite{raffel2020exploring}& \selectfont\color{gray}64.0 & \selectfont\color{gray}71.8 & \selectfont\color{gray}- & \selectfont\color{gray}87.8 & \selectfont\color{gray}48.8 &\selectfont\color{gray} -&\selectfont\color{gray}- &\selectfont\color{gray} -\\
    &\selectfont\color{gray}GIT2 \cite{wang2022git}&\selectfont\color{gray}-&\selectfont\color{gray} - & \selectfont\color{gray}75.8 &\selectfont\color{gray} 145.0&\selectfont\color{gray} -  &\selectfont\color{gray}- &\selectfont\color{gray} -&\selectfont\color{gray}- &\selectfont\color{gray} - \\
    &\selectfont\color{gray} PALI 17B \cite{chen2023pali}& \selectfont\color{gray} mT5-XXL \cite{xue2021mt5}& \selectfont\color{gray} 71.8 &\selectfont\color{gray} 79.9 &\selectfont\color{gray} 160.4 &\selectfont\color{gray} - &\selectfont\color{gray} - &\selectfont\color{gray} -&\selectfont\color{gray}- &\selectfont\color{gray} -\\
     &\selectfont\color{gray}PALI-X 55B \cite{chen2023palix}& \selectfont\color{gray}-& \selectfont\color{gray}80.8 & \selectfont\color{gray}84.5 & \selectfont\color{gray}163.7 & \selectfont\color{gray}86.8 & \selectfont\color{gray}54.8 &\selectfont\color{gray} -&\selectfont\color{gray}- &\selectfont\color{gray} -\\

    &\selectfont\color{gray}PALI 3 \cite{chen2023pali3}&\selectfont\color{gray} UL2 \cite{tay2022unifying}& \selectfont\color{gray}78.3 &\selectfont\color{gray} 85.7 &\selectfont\color{gray} 164.3 &\selectfont\color{gray} 88.6& \selectfont\color{gray} 62.4  &\selectfont\color{gray} -&\selectfont\color{gray}- &\selectfont\color{gray} -\\

    \midrule

    \multirow{12}{*}{\rotatebox{90}{Generalist}} 
    
    & \cellcolor{gray!20}InstructBlip \cite{dai2023instructblip} & \cellcolor{gray!20}Flan-T5-XL \cite{flant5}& \cellcolor{gray!20}64.0 &\cellcolor{gray!20}63.9 & \cellcolor{gray!20}139.9 &\cellcolor{gray!20}77.2 &\cellcolor{gray!20}43.6 &\cellcolor{gray!20}36.3&\cellcolor{gray!20}89.3 & \cellcolor{gray!20}60.4 \\
    &+ \methodname{} & & 67.3 & 66.3 & 145.5 & 85.5 & 51.6 & 40.9 &93.0 & 68.6\\
    
    &$ \Delta $ & & \textcolor{OliveGreen}{+3.3} & \textcolor{OliveGreen}{+2.4} &\textcolor{OliveGreen}{+5.6}& \textcolor{OliveGreen}{+8.3} & \textcolor{OliveGreen}{+8.0} & \textcolor{OliveGreen}{+4.6}  & \textcolor{OliveGreen}{+3.7} & \textcolor{OliveGreen}{+8.2} \\ 
    
     &  \cellcolor{gray!20}InstructBlip  \cite{dai2023instructblip}& \cellcolor{gray!20}Flan-T5-XXL \cite{flant5} &\cellcolor{gray!20}66.8 & \cellcolor{gray!20}65.0 & \cellcolor{gray!20}143.1 &\cellcolor{gray!20}81.1 & \cellcolor{gray!20}49.4 &\cellcolor{gray!20}40.0 & \cellcolor{gray!20}91.6 &\cellcolor{gray!20}65.3 \\
    &+ \methodname{} & & 69.5 & 67.1 & 146.9 & 85.9 &54.2 & 42.9&94.5 & 70.1\\
    &$ \Delta $ & &
    \textcolor{OliveGreen}{+2.7} & \textcolor{OliveGreen}{+2.1} & \textcolor{OliveGreen}{+3.8} & \textcolor{OliveGreen}{+4.8} & \textcolor{OliveGreen}{+4.8} & \textcolor{OliveGreen}{+2.9} & \textcolor{OliveGreen}{+2.9} & \textcolor{OliveGreen}{+4.8} \\ 

    &\cellcolor{gray!20}LLaVA-1.6\cite{liu2024llavanext}&\cellcolor{gray!20}Mistral-7B \cite{jiang2023mistral}&\cellcolor{gray!20}70.6 &\cellcolor{gray!20}71.4 &\cellcolor{gray!20}130.0 &\cellcolor{gray!20}81.2 & \cellcolor{gray!20}46.7 &\cellcolor{gray!20}37.9 & \cellcolor{gray!20}90.6 &\cellcolor{gray!20}64.0 \\
    &+ \methodname{} & & 72.8 & 72.8 & 142.2 & 86.7 & 54.3 & 41.4 & 95.9 & 70.5\\
    &$ \Delta $ & & \textcolor{OliveGreen}{+2.2} & \textcolor{OliveGreen}{+1.4} & \textcolor{OliveGreen}{+12.2} & \textcolor{OliveGreen}{+5.5} & \textcolor{OliveGreen}{+7.6} & \textcolor{OliveGreen}{+3.5} & \textcolor{OliveGreen}{+5.3} & \textcolor{OliveGreen}{+6.5} \\ 

    &  \cellcolor{gray!20}Qwen-VL \cite{bai2023qwenvl}& \cellcolor{gray!20}Qwen-7B \cite{bai2023qwen}& \cellcolor{gray!20}75.6 & \cellcolor{gray!20}77.6 & \cellcolor{gray!20}141.1 & \cellcolor{gray!20}80.5 & \cellcolor{gray!20}40.4 &\cellcolor{gray!20}34.0&\cellcolor{gray!20}98.1&\cellcolor{gray!20}60.5\\ 
    &+ \methodname{} & & 76.4 & 78.8& 141.3 &85.1& 47.8 & 36.7&98.8 & 66.5\\
    &$ \Delta $ & & \textcolor{OliveGreen}{+0.8} & \textcolor{OliveGreen}{+1.2} & \textcolor{OliveGreen}{+0.2} & \textcolor{OliveGreen}{+4.6} & \textcolor{OliveGreen}{+7.4} & \textcolor{OliveGreen}{+2.7} & \textcolor{OliveGreen}{+0.7} & \textcolor{OliveGreen}{+6.0} \\


    \hline
  \end{tabular}
  }
  \vspace{-0.3cm}
  \caption{\textbf{\methodname{} Results}. Quantitative comparison of \methodname{} integrated into InstructBLIP, LLaVA and Qwen-VL  of different sizes, and the corresponding baseline. Evaluation covers scene-text, document understanding benchmarks and 0-shot capabilities, with average performance per domain. \methodname{} consistently outperforms the baselines, demonstrating its effectiveness and versatility}
  \label{tab:results}
  \vspace{-0.3cm}
  
\end{table*}

\begin{figure}[t]
    \centering
  \includegraphics[width=\linewidth]{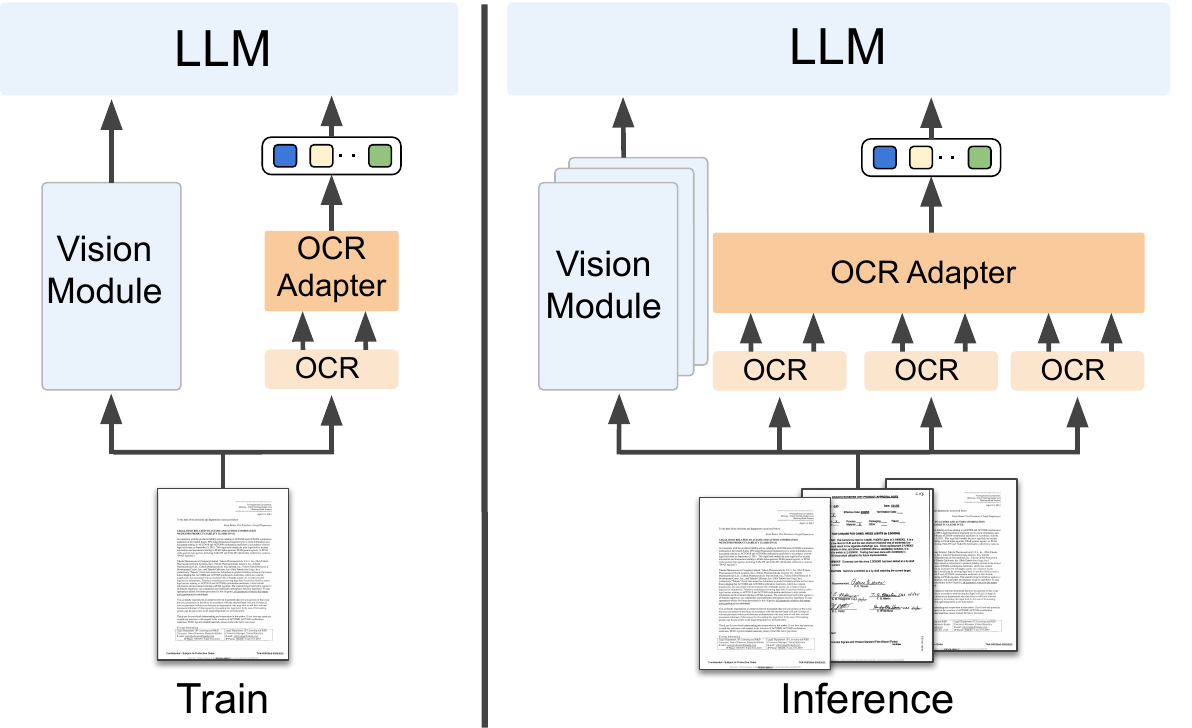}
  \vspace{-0.6cm}
  \caption{\textbf{Extending \methodname{} to multi-page documents}. During inference, we process each page individually through our OCR adapter. Subsequently, we concatenate the results from all pages into a single output, which is then fed into the LLM.}
\label{fig:mp_extension}
  \vspace{-0.3cm}
\end{figure}

\section{Experiments}
\label{sec:exp}
In this section, we demonstrate the advantages of employing \methodname{} to seamlessly integrate OCR information into state-of-the-art VL models in both scene-text and document understanding benchmarks. Initially, we assess various state-of-the-art VL methods, comparing their performance with and without \methodname{} across diverse tasks and domains, including a zero-shot scenario. Subsequently, we highlight the efficacy of incorporating \methodnamelw{}, which creates a rich condensed representation of the OCR, addressing the challenge of long OCR sequences, as in multi-page document question answering.

\subsection{Experimental Setting}
\label{sec:datasets}
For all experiments, we use the open-source versions of the VL architectures and initialize from their weights respectively. Unless mentioned otherwise, the OCR encoder is a based on a T5 large encoder \cite{raffel2020exploring} with 2D layout embedding initially pretrained in a similar fashion to DocFormerv2~\cite{appalaraju2023docformerv2}. We initialize the \qfname{} from the pretrained weights of BERT~\cite{devlin-etal-2019-bert}, the cross attention weights are trained from scratch. To align the \qfname{} with the OCR encoder, we conduct our pretraining protocol on the expansive unlabeled document dataset IDL~\cite{IDL}. The models are fine-tuned on a variety of domains and tasks simultaneously: document question answering, scene-text visual question answering, general visual question answering, scene-text captioning and image captioning . See \cref{supp:implementation_details} for more implementation details.

\subsection{Integrating \methodname{} into leading VL methods}
\label{sec:light}

In \cref{tab:results} we report the results of various VL models on several scene-text and document understanding tasks. In the scene-text domain results are provided for TextVQA and STVQA for question answering and TextCaps for captioning. In the document understanding domain, we utilize DocVQA and InfoVQA. Additionally, we provide zero-shot results on the multipage document understanding dataset DUDE~\cite{dude}. First, we list the specialist models, which were fine-tuned on each dataset independently. These models range from a few hundred million parameters \cite{ganz2023models} to 55B billion parameters \cite{chen2023palix} and are pretrained on different datasets. As such, we report these number only for reference as they are not necessarily comparable.
In the lower section of the table, we present a comparative analysis between our method and baseline approaches, which were trained under identical settings as our method. This comparison includes performance gaps to underscore the benefits achieved. Specifically, we evaluate our approach within the context of VL models such as InstructBLIP \cite{dai2023instructblip}, LLaVA \cite{liu2024llavanext}, and Qwen-VL \cite{bai2023qwenvl}, integrating \methodname{} into each of them. For the InstructBLIP baseline, we consider both the Flan-T5-XL and Flan-T5-XXL architectures \cite{flant5}.

The outcomes of our analyses demonstrate the superior performance of our method compared to the baselines across various architectures and benchmarks. For instance, when integrated with LLaVA, \methodname{} showcases enhancements in TextVQA, DocVQA, and InfoVQA, with notable improvements of \textbf{+2.2\%}, \textbf{+5.5\%}, and \textbf{+7.6\%}, respectively. Furthermore, we calculate the average scores over scene-text and document-based datasets, revealing consistent benefits even when compared to the top-performing method Qwen-VL. Furthermore, we direct the reader to results on non-OCR related benchmarks in \cref{table:vqa_results}, demonstrating comparable or even slightly improved performance compared to the baseline.

\setlength{\tabcolsep}{2pt}
\begin{table}[ht]
\begin{center}
\resizebox{\linewidth}{!}{
\begin{tabular}{lclllc}
\toprule
    \multirow{2}{*}{\textbf{Method}} &  \textbf{TFLOPs} & \multicolumn{3}{c}{\textbf{Datasets}} & \textbf{Average}\\
     &  ($\downarrow$) & DocVQA ($\uparrow$) & InfoVQA ($\uparrow$)& DUDE ($\uparrow$) & Document \\
    \hline & \\[-2ex]
    \rowcolor{gray!20}InstructBlip XL & 3.5 & 77.2 & 43.6 & 36.3 & 60.4 \\
    \textcolor{gray!70}{\hspace{0.3cm}+\methodname{}} & \textcolor{gray!70}{4.4} & \textcolor{gray!70}{85.5 ({+8.3})} & \textcolor{gray!70}{51.6 ({+8.0})} & \textcolor{gray!70}{40.9 ({+4.6})} & \textcolor{gray!70}{68.6} \\
    {\hspace{0.3cm}+\methodnamelw{}} & \textbf{1.3} & 85.4 (\textcolor{OliveGreen}{+8.2}) & 50.4 (\textcolor{OliveGreen}{+6.8}) & 40.1 (\textcolor{OliveGreen}{+3.8}) & 67.9 \\
    \rowcolor{gray!20}InstructBlip XXL & 12.3 & 81.1 & 49.4 & 40.0 & 65.3 \\
    \textcolor{gray!70}{\hspace{0.3cm}+\methodname{}} & \textcolor{gray!70}{13.4} & \textcolor{gray!70}{85.9 ({+4.8})} & \textcolor{gray!70}{54.2 ({+4.8})} & \textcolor{gray!70}{42.9 ({+2.9})} & \textcolor{gray!70}{70.1} \\
    {\hspace{0.3cm}+\methodnamelw{}} & \textbf{1.8} & 84.3 (\textcolor{OliveGreen}{+3.2}) & 51.7 (\textcolor{OliveGreen}{+2.3}) & 41.0 (\textcolor{OliveGreen}{+1.0}) & 68.0 \\
    \rowcolor{gray!20}LLaVA-1.6 & 20.3 & 81.2 & 46.7 & 37.9 & 64.0 \\
    \textcolor{gray!70}{\hspace{0.3cm}+\methodname{}} & \textcolor{gray!70}{21.7} & \textcolor{gray!70}{86.7 ({+5.5})} & \textcolor{gray!70}{54.3 ({+7.6})} & \textcolor{gray!70}{41.4 ({+3.5})} & \textcolor{gray!70}{70.5} \\
    {\hspace{0.3cm}+\methodnamelw{}} & \textbf{5.5} & 84.4 (\textcolor{OliveGreen}{+3.2}) & 51.5 (\textcolor{OliveGreen}{+4.8}) & 40.8 (\textcolor{OliveGreen}{+2.9}) & 68.0 \\
    \rowcolor{gray!20}Qwen-VL & 19.8 & 80.5 & 40.4 & 34.0 & 60.5 \\
    \textcolor{gray!70}{\hspace{0.3cm}+\methodname{}} & \textcolor{gray!70}{21.0} & \textcolor{gray!70}{85.1 ({+4.6})} & \textcolor{gray!70}{47.8 ({+7.4})} & \textcolor{gray!70}{36.7 ({+2.7})} & \textcolor{gray!70}{66.5} \\
    {\hspace{0.3cm}+\methodnamelw{}} & \textbf{5.4} & 85.1 (\textcolor{OliveGreen}{+4.6}) & 48.2 (\textcolor{OliveGreen}{+7.8}) & 38.8 (\textcolor{OliveGreen}{+4.8}) & 66.6 \\
\bottomrule
\end{tabular}}
\vspace{-0.2cm}
\caption{\textbf{\methodnamelw{} Results}. Outcomes of integrating \methodnamelw{} into the base model, with \methodname{} results for comparison. Alongside performance metrics, we provide the TFLOPs. Our compression technique reduces computational costs while maintaining improved performance.}
\label{tab:tap_vl_light}
\setlength{\tabcolsep}{1.4pt}
\end{center}
\vspace{-0.9cm}
\end{table}

In the realm of document analysis, significant improvements are observed, with average performance gaps reaching up to \textbf{8.2\%}. While \methodname{} demonstrates improvements in both scene-text and document datasets, it can be seen that in the document datasets the improvements are larger. This suggests that in the documents domain, a dedicated OCR encoder has a pivotal role. This aligns with the typically dense textual content and intricate structures found in documents. Consequently, a dedicated OCR encoder, equipped with spatial information as input, possesses the capability to provide more comprehensive answers compared to the VL model alone, which is confronted with raw, unstructured text. Therefore, the effective integration of this essential component through our method equips the VL model with the supplementary information required to address the tasks proficiently.

\begin{figure*}[t!]
    \centering
  \includegraphics[width=\linewidth]{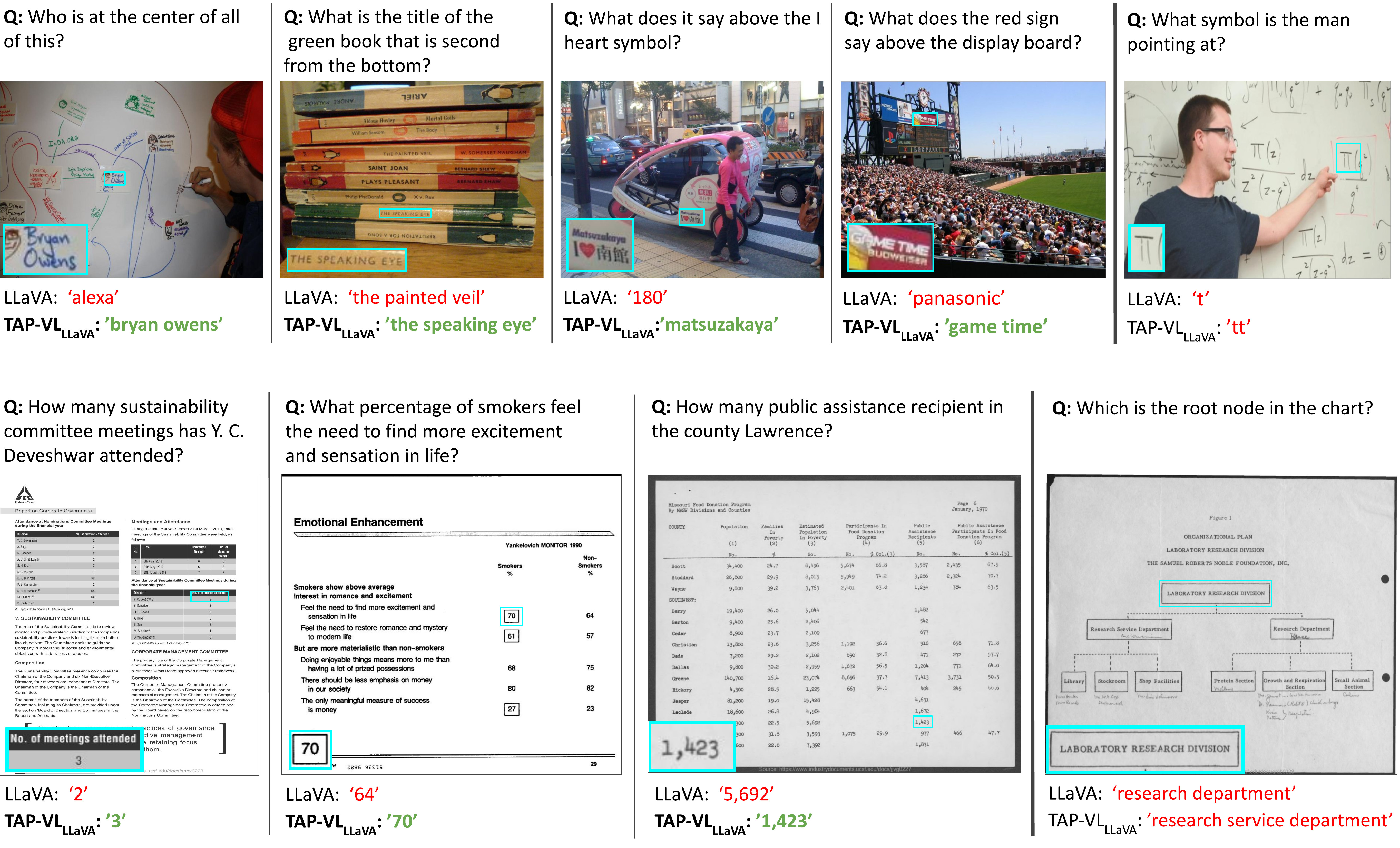}
  \vspace{-0.6cm}
  \caption{\textbf{Qualitative Improvements Demonstrated by \methodname{}}. Illustrative examples showcasing the improvements achieved by our method on scene-text (top) and document understanding (bottom) benchmarks using LLaVA. \methodname{} enhances the base model's ability to grasp OCR and layout information, yielding significant improvements across both benchmark types.}
\label{fig:qualitative_examples}
  \vspace{-0.6cm}
\end{figure*}

To evaluate the performance in a zero-shot setting, we opt for the multi-page DocVQA scenario. Particularly, we combine the OCR from each page to create a single long OCR sequence representing the entire document. The visual encoding is applied individually to each page and subsequently concatenated before being fed into the VL model's LLM (see Figure \ref{fig:mp_extension}). As demonstrated in the \cref{tab:results}, our method yields significant improvements in the multi-page zero-shot scenario (up to \textbf{4.6\%} improvement).

After presenting the quantitative impact of using \methodname{}, we now introduce qualitative findings. In Figure \ref{fig:qualitative_examples}, we display how our method integrates into LLaVA-1.6. The top row shows examples from scene-text benchmarks, while the bottom row features document-related benchmarks. Our method notably enhances the VL model's OCR and layout comprehension, leading to improved performance. For instance, in the second top-row example, the base model struggles to identify the "\textit{second from the bottom}" book, whereas \methodname{} effectively uses layout information to understand it. Similarly, in the bottom row, our model shows significant improvements in comprehending complex structures like tables. 

\textbf{Leveraging \methodname{} for efficiency} 
In this section we present results on \methodnamelw{}, a light-weight version of \methodname{} that improves OCR-oriented tasks while reducing computational load. Processing extended sequences demands substantial computational resources, this is especially significant in transformers which are composed of attention blocks that require quadratic complexity computation. \methodnamelw{} utilizes the condensed representation alone, omitting raw text OCR input. This is particularly crucial for documents, which consist of dense-text images. In \methodnamelw{}, the sequence length of the input to the LLM remains unaffected by the length of the OCR, and is bounded to $K$ tokens.  \cref{tab:tap_vl_light} shows \methodnamelw{} results on document understanding tasks.

Our method improves performance across all benchmarks while offering computational advantages. The DUDE benchmark, known for its multi-page document VQA datasets containing sequences that extend up to 10K tokens, poses a significant challenge to VL systems, pushing them to their operational limits. For example, when examining the integration of \methodname{} into LLaVA, our compressed OCR version yields improvements of \textbf{+3.2\%}, \textbf{+4.8\%}, and \textbf{+2.9\%} on DocVQA, InfoVQA, and DUDE, respectively, while reducing TFLOPs from 20.3 to 5.5. Moreover, in the case of Qwen-VL, the benefits of inputting the raw OCR sequence into the LLM are negligible. 

\begin{table}[t]
\begin{center}
  \vspace{-0.2cm}
\resizebox{\linewidth}{!}{
\begin{tabular}{lcccccc}
\toprule
    \multirow{3}{*}{\textbf{Method}} & \multicolumn{4}{c}{\textbf{\methodname{}'s Components}} & \multicolumn{2}{c}{\textbf{Average}}\\
     & Layout      & OCR- & \multirow{2}{*}{Pretraining} & OCR-to-language &  \multirow{2}{*}{Scene-Text} &  \multirow{2}{*}{Documents} \\
     & information & Module & & alignment &  &  \\
    \hline & \\[-2ex]
    InstructBlip & \xmark & \xmark & \xmark & \xmark & 89.3 & 60.4 \\
    \hdashline & \\[-2ex]
     & 2D-embeddings & \xmark & \xmark & \xmark &  90.0 & 60.0 \\
     \hdashline & \\[-2ex]
     & \multirow{4}{*}{OCR-encoder} 
     & \cmark & \xmark & \xmark & 87.6 & 61.2 \\
     &  & \cmark & \xmark & \cmark & 89.1 & 60.5 \\
    &  & \cmark & \cmark & \xmark  & 89.0 & 60.3 \\
    \rowcolor{gray!20}\methodname{} &  & \cmark & \cmark & \cmark & 92.0  & 65.7 \\
\bottomrule
\end{tabular}}
\vspace{-0.3cm}
\caption{\textbf{Analysis of \methodname{}'s Components through Ablation Studies}. We examine the impact of each component of \methodname{}. Interestingly, only when \qfname{} is combined with our layout-aware pretraining and OCR-to-language alignment, our method yields consistent performance improvements.}
\setlength{\tabcolsep}{1.4pt}
\label{table:ablations_blop_components}
\end{center}
  \vspace{-0.8cm}
\end{table}

\section{Ablation studies}

This section analyzes performance improvements from our proposed architecture and pretraining framework. We examine the effects of \methodname{}'s individual components, pretraining objectives, and data quantity on the InstructBlip (Flan-T5-XL) architecture with T5 base OCR encoder.

\textbf{\methodname{} components:}
In \cref{table:ablations_blop_components}, we incrementally incorporate \methodname{}'s components to assess their individual effects, starting from the InstructBlip baseline. Initially, we explore a naive method to integrate layout information into OCR by adding a 2D embedding for each OCR token derived from its spatial location via a designated spatial embedding layer in a residual manner. This methods perform similarly to the baseline, suggesting incomplete utilization of residual or encoded information. Subsequently, integrating our OCR module without additional pretraining improves document results by 0.8\% but decreases scene-text results by 1.7\%. Further, we analyze the impact of our pretraining and OCR-to-language alignment step. While applying each separately leads to improvements in scene-text and degradation in documents, applying both steps results in consistent enhancement in both domains.

\textbf{Pretraining Objective:}
We conduct an ablation study focusing on different configurations of our pretraining tasks. We incrementally incorporate the three pretraining objectives: OCR-Grounded Mask Denoising, OCR Mask Contrastive and OCR-Grounded Mask Matching. As indicated in \cref{table:ablations_pt_tasks_and_data}, each pretraining task contributes to enhancing the overall effectiveness of the final model. 

\setlength{\tabcolsep}{4pt}
\begin{table}[t]
\begin{center}
\resizebox{0.45\textwidth}{!}{
\begin{tabular}{ccccccc}
\toprule
    \multicolumn{3}{c}{\textbf{Pretraining Tasks}} & \multirow{2}{*}{\textbf{Samples}} & \multicolumn{2}{c}{\textbf{Average}}\\
     Denoising & Contrastive & Matching & & Scene-Text & Documents \\
    \hline & \\[-2ex]
    \xmark & \xmark & \xmark & - & 86.8\textsuperscript{$\dagger$} & 58.3\textsuperscript{$\dagger$} \\
    \cmark & \xmark & \xmark & 13M & 85.9\textsuperscript{$\dagger$} & 60.7\textsuperscript{$\dagger$} \\
    \cmark & \cmark & \xmark & 13M & 87.5\textsuperscript{$\dagger$} & 62.0\textsuperscript{$\dagger$} \\
    \rowcolor{gray!20} \cmark & \cmark & \cmark & 13M & 89.5\textsuperscript{$\dagger$} & 62.2\textsuperscript{$\dagger$} \\
    \cdashline{1-6} & \\[-2ex]
    \cmark & \cmark & \cmark & 2M & 91.2 & 62.8 \\
    \cmark & \cmark & \cmark & 13M & 91.7 & 63.4 \\
    \cmark & \cmark & \cmark & 22M & 91.0 & 65.2 \\
    \rowcolor{gray!20} \cmark & \cmark & \cmark & 31M & 92.0 & 65.7 \\
\bottomrule
\end{tabular}
}
\vspace{-0.2cm}
\caption{\textbf{Pretraining Objectives and Data}. The top section of the table depicts the influence of each pretraining objective. Below, we investigate the impact of increasing the data size and pretraining iterations. Notably, performance in both document and scene-text tasks shows ongoing improvement without reaching a plateau, even after processing 31 million documents. \textsuperscript{$\dagger$} Experiments which underwent shorter training.}
\label{table:ablations_pt_tasks_and_data}
\setlength{\tabcolsep}{1.4pt}
\end{center}
\vspace{-0.9cm}
\end{table}

\textbf{Scale of pretraining Data:}
We examine the impact of varying data volumes during pretraining, as shown in \cref{table:ablations_pt_tasks_and_data}. The results reveal a correlation between data volume and model performance on both scene-text and document benchmarks. This relationship is particularly notable in the case of document benchmarks, where results exhibit an upward trend with increased pretraining length, rising from 62.8 with 2M pretraining samples to 65.7 with 31M training samples. While the correlation persists in scene-text benchmarks, we note a decline in performance when pretraining samples increase from 13M to 22M. However, the best result is achieved with the maximum pretraining samples, 31M, indicating that pretraining on documents has the potential to enhance OCR-oriented scene-text performance.

\textbf{Performance on General Benchmarks:} We evaluated our system on a general VQA dataset (VQAv2) and a CAPS dataset (COCO), which do not specifically require OCR. Our analysis indicates that \methodname{} either preserves or enhances the non-OCR capabilities of the baseline vision-language model. Integration with InstructBlip (XXL) resulted in a performance boost on the COCO dataset by 3.1\%, with only a marginal decline on VQAv2 of 0.1\%.

\begin{table}[t!]
  \centering
  
  \label{tab:general_results}
            \vspace{-0.2cm}

  \begin{tabular}{ll|ccc cc c | cc }
    \midrule
    
    \textbf{Method} & \textbf{LLM} &  VQAv2 & COCO \\
    
    & & \fontsize{7pt}{8pt}\selectfont VQAScore & \fontsize{7pt}{8pt}\selectfont CIDEr \\

    \cline{1-4} &  & & & &\\[-2ex] 
   
    
     \cellcolor{gray!20}InstructBlip \cite{dai2023instructblip} & \cellcolor{gray!20}Flan-T5-XL & \cellcolor{gray!20}78.2 &\cellcolor{gray!20}141.1 \\
    + \methodnamelw{} & & 78.2 & 143.1 \\
    + \methodname{} & & 78.1 & 143.8 \\
    
    
       \cellcolor{gray!20}InstructBlip  \cite{dai2023instructblip}& \cellcolor{gray!20}Flan-T5-XXL &\cellcolor{gray!20}78.9 & \cellcolor{gray!20}138.2 \\
    + \methodnamelw{} & & 78.5 & 141.3 
    \\
    + \methodname{} & & 78.8 & 141.3 \\

    \cellcolor{gray!20}LLaVA-1.6 \cite{liu2024llavanext}&\cellcolor{gray!20}Mistral-7B &\cellcolor{gray!20}80.7 &\cellcolor{gray!20}133.3 \\
    + \methodnamelw{} & & 80.7 & 135.3 \\
    + \methodname{} & & 80.8 & 135.5 \\
    \cellcolor{gray!20}Qwen-VL \cite{liu2023improved}& \cellcolor{gray!20}Qwen-7B & \cellcolor{gray!20}80.4 & \cellcolor{gray!20}126.7 \\ 
    + \methodnamelw{} & & 80.4&127.2\\
    + \methodname{} & & 80.3&127.2\\


    \hline
  \end{tabular}
  \vspace{-0.3cm}
  \caption{\textbf{\methodname{} Results}. \methodname{} results on VQAv2 and COCO, integrated into InstructBLIP, LLaVA-1.6, and Qwen-VL.}
  \vspace{-0.57cm}
  \label{table:vqa_results}
\end{table}


\vspace{-0.1cm}
\section{Discussion and Conclusion}
\vspace{-0.1cm}
In this study, we proposed \methodname{}, a novel method for integrating OCR information into VL models. Our approach, which treats OCR as a distinct modality, utilizes a lightweight transformer-based OCR adapter to compress OCR and layout information into fixed-length sequences for input into VL models. Through extensive experiments, we demonstrated consistent performance improvements across various benchmarks, including both natural images and document-based VL tasks. Additionally, we proposed \methodnamelw{} which significantly reduces computational costs by using a concise representation of the OCR. Overall, our findings suggest that integrating OCR information into VL models can lead to substantial performance gains and computational savings, making it a promising avenue for future research in the field.

{
    \small
    \bibliographystyle{ieeenat_fullname}
    \bibliography{egbib}

\begin{thebibliography}{57}
\providecommand{\natexlab}[1]{#1}
\providecommand{\url}[1]{\texttt{#1}}
\expandafter\ifx\csname urlstyle\endcsname\relax
  \providecommand{\doi}[1]{doi: #1}\else
  \providecommand{\doi}{doi: \begingroup \urlstyle{rm}\Url}\fi

\bibitem[Aberdam et~al.(2021)Aberdam, Litman, Tsiper, Anschel, Slossberg, Mazor, Manmatha, and Perona]{aberdam2021sequence}
Aviad Aberdam, Ron Litman, Shahar Tsiper, Oron Anschel, Ron Slossberg, Shai Mazor, R Manmatha, and Pietro Perona.
\newblock Sequence-to-sequence contrastive learning for text recognition.
\newblock In \emph{Proceedings of the IEEE/CVF Conference on Computer Vision and Pattern Recognition}, pages 15302--15312, 2021.

\bibitem[Aberdam et~al.(2022)Aberdam, Ganz, Mazor, and Litman]{aberdam2022multimodal}
Aviad Aberdam, Roy Ganz, Shai Mazor, and Ron Litman.
\newblock Multimodal semi-supervised learning for text recognition.
\newblock \emph{arXiv preprint arXiv:2205.03873}, 2022.

\bibitem[Aberdam et~al.(2023)Aberdam, Bensa{\"\i}d, Golts, Ganz, Nuriel, Tichauer, Mazor, and Litman]{aberdam2023clipter}
Aviad Aberdam, David Bensa{\"\i}d, Alona Golts, Roy Ganz, Oren Nuriel, Royee Tichauer, Shai Mazor, and Ron Litman.
\newblock Clipter: Looking at the bigger picture in scene text recognition.
\newblock In \emph{Proceedings of the IEEE/CVF International Conference on Computer Vision}, pages 21706--21717, 2023.

\bibitem[Alayrac et~al.(2022)Alayrac, Donahue, Luc, Miech, Barr, Hasson, Lenc, Mensch, Millican, Reynolds, et~al.]{alayrac2022flamingo}
Jean-Baptiste Alayrac, Jeff Donahue, Pauline Luc, Antoine Miech, Iain Barr, Yana Hasson, Karel Lenc, Arthur Mensch, Katherine Millican, Malcolm Reynolds, et~al.
\newblock Flamingo: a visual language model for few-shot learning.
\newblock \emph{Advances in Neural Information Processing Systems}, 35:\penalty0 23716--23736, 2022.

\bibitem[Antol et~al.(2015)Antol, Agrawal, Lu, Mitchell, Batra, Zitnick, and Parikh]{antol2015vqa}
Stanislaw Antol, Aishwarya Agrawal, Jiasen Lu, Margaret Mitchell, Dhruv Batra, C~Lawrence Zitnick, and Devi Parikh.
\newblock Vqa: Visual question answering.
\newblock In \emph{Proceedings of the IEEE international conference on computer vision}, pages 2425--2433, 2015.

\bibitem[Appalaraju et~al.(2023)Appalaraju, Tang, Dong, Sankaran, Zhou, and Manmatha]{appalaraju2023docformerv2}
Srikar Appalaraju, Peng Tang, Qi Dong, Nishant Sankaran, Yichu Zhou, and R. Manmatha.
\newblock Docformerv2: Local features for document understanding, 2023.

\bibitem[Bai et~al.(2023{\natexlab{a}})Bai, Bai, Chu, Cui, Dang, Deng, Fan, Ge, Han, Huang, Hui, Ji, Li, Lin, Lin, Liu, Liu, Lu, Lu, Ma, Men, Ren, Ren, Tan, Tan, Tu, Wang, Wang, Wang, Wu, Xu, Xu, Yang, Yang, Yang, Yang, Yao, Yu, Yuan, Yuan, Zhang, Zhang, Zhang, Zhang, Zhou, Zhou, Zhou, and Zhu]{bai2023qwen}
Jinze Bai, Shuai Bai, Yunfei Chu, Zeyu Cui, Kai Dang, Xiaodong Deng, Yang Fan, Wenbin Ge, Yu Han, Fei Huang, Binyuan Hui, Luo Ji, Mei Li, Junyang Lin, Runji Lin, Dayiheng Liu, Gao Liu, Chengqiang Lu, Keming Lu, Jianxin Ma, Rui Men, Xingzhang Ren, Xuancheng Ren, Chuanqi Tan, Sinan Tan, Jianhong Tu, Peng Wang, Shijie Wang, Wei Wang, Shengguang Wu, Benfeng Xu, Jin Xu, An Yang, Hao Yang, Jian Yang, Shusheng Yang, Yang Yao, Bowen Yu, Hongyi Yuan, Zheng Yuan, Jianwei Zhang, Xingxuan Zhang, Yichang Zhang, Zhenru Zhang, Chang Zhou, Jingren Zhou, Xiaohuan Zhou, and Tianhang Zhu.
\newblock Qwen technical report, 2023{\natexlab{a}}.

\bibitem[Bai et~al.(2023{\natexlab{b}})Bai, Bai, Yang, Wang, Tan, Wang, Lin, Zhou, and Zhou]{bai2023qwenvl}
Jinze Bai, Shuai Bai, Shusheng Yang, Shijie Wang, Sinan Tan, Peng Wang, Junyang Lin, Chang Zhou, and Jingren Zhou.
\newblock Qwen-vl: A versatile vision-language model for understanding, localization, text reading, and beyond, 2023{\natexlab{b}}.

\bibitem[Biten et~al.(2019)Biten, Tito, Mafla, Gomez, Rusiñol, Jawahar, Valveny, and Karatzas]{stvqa}
Ali~Furkan Biten, Rubèn Tito, Andrés Mafla, Lluis Gomez, Marçal Rusiñol, C.V. Jawahar, Ernest Valveny, and Dimosthenis Karatzas.
\newblock Scene text visual question answering.
\newblock In \emph{2019 IEEE/CVF International Conference on Computer Vision (ICCV)}, pages 4290--4300, 2019.

\bibitem[Biten et~al.(2022)Biten, Litman, Xie, Appalaraju, and Manmatha]{latr}
Ali~Furkan Biten, Ron Litman, Yusheng Xie, Srikar Appalaraju, and R. Manmatha.
\newblock Latr: Layout-aware transformer for scene-text vqa.
\newblock In \emph{2022 IEEE/CVF Conference on Computer Vision and Pattern Recognition (CVPR)}, pages 16527--16537, 2022.

\bibitem[Biten et~al.(2023)Biten, Tito, Gomez, Valveny, and Karatzas]{IDL}
Ali~Furkan Biten, Rub{\`e}n Tito, Lluis Gomez, Ernest Valveny, and Dimosthenis Karatzas.
\newblock Ocr-idl: Ocr annotations for industry document library dataset.
\newblock In \emph{Computer Vision -- ECCV 2022 Workshops}, pages 241--252, Cham, 2023. Springer Nature Switzerland.

\bibitem[Blau et~al.(2024)Blau, Fogel, Ronen, Golts, Ganz, Ben~Avraham, Aberdam, Tsiper, and Litman]{blau2024gram}
Tsachi Blau, Sharon Fogel, Roi Ronen, Alona Golts, Roy Ganz, Elad Ben~Avraham, Aviad Aberdam, Shahar Tsiper, and Ron Litman.
\newblock Gram: Global reasoning for multi-page vqa.
\newblock In \emph{Proceedings of the IEEE/CVF Conference on Computer Vision and Pattern Recognition}, pages 15598--15607, 2024.

\bibitem[Chen et~al.(2023{\natexlab{a}})Chen, Han, Zhao, Zhang, Shi, Xu, and Xu]{chen2023xllm}
Feilong Chen, Minglun Han, Haozhi Zhao, Qingyang Zhang, Jing Shi, Shuang Xu, and Bo Xu.
\newblock X-llm: Bootstrapping advanced large language models by treating multi-modalities as foreign languages, 2023{\natexlab{a}}.

\bibitem[Chen et~al.(2020)Chen, Kornblith, Norouzi, and Hinton]{chen2020simple}
Ting Chen, Simon Kornblith, Mohammad Norouzi, and Geoffrey Hinton.
\newblock A simple framework for contrastive learning of visual representations, 2020.

\bibitem[Chen et~al.(2015)Chen, Fang, Lin, Vedantam, Gupta, Dollar, and Zitnick]{chen2015microsoft}
Xinlei Chen, Hao Fang, Tsung-Yi Lin, Ramakrishna Vedantam, Saurabh Gupta, Piotr Dollar, and C.~Lawrence Zitnick.
\newblock Microsoft coco captions: Data collection and evaluation server, 2015.

\bibitem[Chen et~al.(2023{\natexlab{b}})Chen, Djolonga, Padlewski, Mustafa, Changpinyo, Wu, Ruiz, Goodman, Wang, Tay, Shakeri, Dehghani, Salz, Lucic, Tschannen, Nagrani, Hu, Joshi, Pang, Montgomery, Pietrzyk, Ritter, Piergiovanni, Minderer, Pavetic, Waters, Li, Alabdulmohsin, Beyer, Amelot, Lee, Steiner, Li, Keysers, Arnab, Xu, Rong, Kolesnikov, Seyedhosseini, Angelova, Zhai, Houlsby, and Soricut]{chen2023palix}
Xi Chen, Josip Djolonga, Piotr Padlewski, Basil Mustafa, Soravit Changpinyo, Jialin Wu, Carlos~Riquelme Ruiz, Sebastian Goodman, Xiao Wang, Yi Tay, Siamak Shakeri, Mostafa Dehghani, Daniel Salz, Mario Lucic, Michael Tschannen, Arsha Nagrani, Hexiang Hu, Mandar Joshi, Bo Pang, Ceslee Montgomery, Paulina Pietrzyk, Marvin Ritter, AJ Piergiovanni, Matthias Minderer, Filip Pavetic, Austin Waters, Gang Li, Ibrahim Alabdulmohsin, Lucas Beyer, Julien Amelot, Kenton Lee, Andreas~Peter Steiner, Yang Li, Daniel Keysers, Anurag Arnab, Yuanzhong Xu, Keran Rong, Alexander Kolesnikov, Mojtaba Seyedhosseini, Anelia Angelova, Xiaohua Zhai, Neil Houlsby, and Radu Soricut.
\newblock Pali-x: On scaling up a multilingual vision and language model, 2023{\natexlab{b}}.

\bibitem[Chen et~al.(2023{\natexlab{c}})Chen, Wang, Beyer, Kolesnikov, Wu, Voigtlaender, Mustafa, Goodman, Alabdulmohsin, Padlewski, Salz, Xiong, Vlasic, Pavetic, Rong, Yu, Keysers, Zhai, and Soricut]{chen2023pali3}
Xi Chen, Xiao Wang, Lucas Beyer, Alexander Kolesnikov, Jialin Wu, Paul Voigtlaender, Basil Mustafa, Sebastian Goodman, Ibrahim Alabdulmohsin, Piotr Padlewski, Daniel Salz, Xi Xiong, Daniel Vlasic, Filip Pavetic, Keran Rong, Tianli Yu, Daniel Keysers, Xiaohua Zhai, and Radu Soricut.
\newblock Pali-3 vision language models: Smaller, faster, stronger, 2023{\natexlab{c}}.

\bibitem[Chen et~al.(2023{\natexlab{d}})Chen, Wang, Changpinyo, Piergiovanni, Padlewski, Salz, Goodman, Grycner, Mustafa, Beyer, Kolesnikov, Puigcerver, Ding, Rong, Akbari, Mishra, Xue, Thapliyal, Bradbury, Kuo, Seyedhosseini, Jia, Ayan, Riquelme, Steiner, Angelova, Zhai, Houlsby, and Soricut]{chen2023pali}
Xi Chen, Xiao Wang, Soravit Changpinyo, AJ Piergiovanni, Piotr Padlewski, Daniel Salz, Sebastian Goodman, Adam Grycner, Basil Mustafa, Lucas Beyer, Alexander Kolesnikov, Joan Puigcerver, Nan Ding, Keran Rong, Hassan Akbari, Gaurav Mishra, Linting Xue, Ashish Thapliyal, James Bradbury, Weicheng Kuo, Mojtaba Seyedhosseini, Chao Jia, Burcu~Karagol Ayan, Carlos Riquelme, Andreas Steiner, Anelia Angelova, Xiaohua Zhai, Neil Houlsby, and Radu Soricut.
\newblock Pali: A jointly-scaled multilingual language-image model, 2023{\natexlab{d}}.

\bibitem[Chung et~al.(2022)Chung, Hou, Longpre, Zoph, Tay, Fedus, Li, Wang, Dehghani, Brahma, Webson, Gu, Dai, Suzgun, Chen, Chowdhery, Castro-Ros, Pellat, Robinson, Valter, Narang, Mishra, Yu, Zhao, Huang, Dai, Yu, Petrov, Chi, Dean, Devlin, Roberts, Zhou, Le, and Wei]{flant5}
Hyung~Won Chung, Le Hou, Shayne Longpre, Barret Zoph, Yi Tay, William Fedus, Yunxuan Li, Xuezhi Wang, Mostafa Dehghani, Siddhartha Brahma, Albert Webson, Shixiang~Shane Gu, Zhuyun Dai, Mirac Suzgun, Xinyun Chen, Aakanksha Chowdhery, Alex Castro-Ros, Marie Pellat, Kevin Robinson, Dasha Valter, Sharan Narang, Gaurav Mishra, Adams Yu, Vincent Zhao, Yanping Huang, Andrew Dai, Hongkun Yu, Slav Petrov, Ed~H. Chi, Jeff Dean, Jacob Devlin, Adam Roberts, Denny Zhou, Quoc~V. Le, and Jason Wei.
\newblock Scaling instruction-finetuned language models, 2022.

\bibitem[Dai et~al.(2023)Dai, Li, Li, Tiong, Zhao, Wang, Li, Fung, and Hoi]{dai2023instructblip}
Wenliang Dai, Junnan Li, Dongxu Li, Anthony Meng~Huat Tiong, Junqi Zhao, Weisheng Wang, Boyang Li, Pascale Fung, and Steven Hoi.
\newblock Instructblip: Towards general-purpose vision-language models with instruction tuning, 2023.

\bibitem[Devlin et~al.(2019)Devlin, Chang, Lee, and Toutanova]{devlin-etal-2019-bert}
Jacob Devlin, Ming-Wei Chang, Kenton Lee, and Kristina Toutanova.
\newblock {BERT}: Pre-training of deep bidirectional transformers for language understanding.
\newblock In \emph{Proceedings of the 2019 Conference of the North {A}merican Chapter of the Association for Computational Linguistics: Human Language Technologies, Volume 1 (Long and Short Papers)}, pages 4171--4186, Minneapolis, Minnesota, 2019. Association for Computational Linguistics.

\bibitem[Dong et~al.(2019)Dong, Yang, Wang, Wei, Liu, Wang, Gao, Zhou, and Hon]{dong2019unified}
Li Dong, Nan Yang, Wenhui Wang, Furu Wei, Xiaodong Liu, Yu Wang, Jianfeng Gao, Ming Zhou, and Hsiao-Wuen Hon.
\newblock Unified language model pre-training for natural language understanding and generation.
\newblock \emph{Advances in neural information processing systems}, 32, 2019.

\bibitem[Ganz and Elad(2024)]{ganz2024clipag}
Roy Ganz and Michael Elad.
\newblock Clipag: Towards generator-free text-to-image generation.
\newblock In \emph{Proceedings of the IEEE/CVF Winter Conference on Applications of Computer Vision}, pages 3843--3853, 2024.

\bibitem[Ganz et~al.(2023)Ganz, Nuriel, Aberdam, Kittenplon, Mazor, and Litman]{ganz2023models}
Roy Ganz, Oren Nuriel, Aviad Aberdam, Yair Kittenplon, Shai Mazor, and Ron Litman.
\newblock Towards models that can see and read.
\newblock In \emph{Proceedings of the IEEE/CVF international conference on computer vision}, pages 21718--21728, 2023.

\bibitem[Ganz et~al.(2024)Ganz, Kittenplon, Aberdam, Avraham, Nuriel, Mazor, and Litman]{ganz2024question}
Roy Ganz, Yair Kittenplon, Aviad Aberdam, Elad~Ben Avraham, Oren Nuriel, Shai Mazor, and Ron Litman.
\newblock Question aware vision transformer for multimodal reasoning.
\newblock \emph{arXiv preprint arXiv:2402.05472}, 2024.

\bibitem[Goyal et~al.(2017)Goyal, Khot, Summers-Stay, Batra, and Parikh]{vqav2}
Yash Goyal, Tejas Khot, Douglas Summers-Stay, Dhruv Batra, and Devi Parikh.
\newblock Making the v in vqa matter: Elevating the role of image understanding in visual question answering.
\newblock In \emph{Proceedings of the IEEE Conference on Computer Vision and Pattern Recognition (CVPR)}, 2017.

\bibitem[Hsu et~al.(2024)Hsu, Liu, Li, Fujinuma, Nadejde, Niu, Kittenplon, Litman, and Pappagari]{hsu2024m3t}
Benjamin Hsu, Xiaoyu Liu, Huayang Li, Yoshinari Fujinuma, Maria Nadejde, Xing Niu, Yair Kittenplon, Ron Litman, and Raghavendra Pappagari.
\newblock M3t: A new benchmark dataset for multi-modal document-level machine translation.
\newblock \emph{arXiv preprint arXiv:2406.08255}, 2024.

\bibitem[Hu et~al.(2021)Hu, Shen, Wallis, Allen-Zhu, Li, Wang, Wang, and Chen]{hu2021lora}
Edward~J. Hu, Yelong Shen, Phillip Wallis, Zeyuan Allen-Zhu, Yuanzhi Li, Shean Wang, Lu Wang, and Weizhu Chen.
\newblock Lora: Low-rank adaptation of large language models, 2021.

\bibitem[Hu et~al.(2023)Hu, Xu, Li, Li, Chen, and Tu]{hu2023bliva}
Wenbo Hu, Yifan Xu, Yi Li, Weiyue Li, Zeyuan Chen, and Zhuowen Tu.
\newblock Bliva: A simple multimodal llm for better handling of text-rich visual questions.
\newblock \emph{arXiv preprint arXiv:2308.09936}, 2023.

\bibitem[Jiang et~al.(2023)Jiang, Sablayrolles, Mensch, Bamford, Chaplot, de~las Casas, Bressand, Lengyel, Lample, Saulnier, Lavaud, Lachaux, Stock, Scao, Lavril, Wang, Lacroix, and Sayed]{jiang2023mistral}
Albert~Q. Jiang, Alexandre Sablayrolles, Arthur Mensch, Chris Bamford, Devendra~Singh Chaplot, Diego de~las Casas, Florian Bressand, Gianna Lengyel, Guillaume Lample, Lucile Saulnier, Lélio~Renard Lavaud, Marie-Anne Lachaux, Pierre Stock, Teven~Le Scao, Thibaut Lavril, Thomas Wang, Timothée Lacroix, and William~El Sayed.
\newblock Mistral 7b, 2023.

\bibitem[Landeghem et~al.(2023)Landeghem, Tito, Łukasz Borchmann, Pietruszka, Józiak, Powalski, Jurkiewicz, Coustaty, Ackaert, Valveny, Blaschko, Moens, and Stanisławek]{dude}
Jordy~Van Landeghem, Rubén Tito, Łukasz Borchmann, Michał Pietruszka, Paweł Józiak, Rafał Powalski, Dawid Jurkiewicz, Mickaël Coustaty, Bertrand Ackaert, Ernest Valveny, Matthew Blaschko, Sien Moens, and Tomasz Stanisławek.
\newblock Document understanding dataset and evaluation (dude), 2023.

\bibitem[Li et~al.(2021)Li, Selvaraju, Gotmare, Joty, Xiong, and Hoi]{li2021align}
Junnan Li, Ramprasaath Selvaraju, Akhilesh Gotmare, Shafiq Joty, Caiming Xiong, and Steven Chu~Hong Hoi.
\newblock Align before fuse: Vision and language representation learning with momentum distillation.
\newblock \emph{Advances in neural information processing systems}, 34:\penalty0 9694--9705, 2021.

\bibitem[Li et~al.(2022)Li, Li, Xiong, and Hoi]{li2022blip}
Junnan Li, Dongxu Li, Caiming Xiong, and Steven Hoi.
\newblock Blip: Bootstrapping language-image pre-training for unified vision-language understanding and generation.
\newblock In \emph{International Conference on Machine Learning}, pages 12888--12900. PMLR, 2022.

\bibitem[Li et~al.(2023)Li, Li, Savarese, and Hoi]{Li2023BLIP2BL}
Junnan Li, Dongxu Li, Silvio Savarese, and Steven C.~H. Hoi.
\newblock Blip-2: Bootstrapping language-image pre-training with frozen image encoders and large language models.
\newblock In \emph{International Conference on Machine Learning}, 2023.

\bibitem[Litman et~al.()Litman, Tsiper, Tichauer, Appalaraju, Mazor, and Manmatha]{litmanvisfocus}
Ron Litman, Shahar Tsiper, Royee Tichauer, Srikar Appalaraju, Shai Mazor, and R Manmatha.
\newblock Visfocus: Prompt-guided vision encoders for ocr-free dense document understanding.

\bibitem[Litman et~al.(2020)Litman, Anschel, Tsiper, Litman, Mazor, and Manmatha]{litman2020scatter}
Ron Litman, Oron Anschel, Shahar Tsiper, Roee Litman, Shai Mazor, and R Manmatha.
\newblock Scatter: selective context attentional scene text recognizer.
\newblock In \emph{proceedings of the IEEE/CVF conference on computer vision and pattern recognition}, pages 11962--11972, 2020.

\bibitem[Liu et~al.(2023{\natexlab{a}})Liu, Li, Li, and Lee]{liu2023improved}
Haotian Liu, Chunyuan Li, Yuheng Li, and Yong~Jae Lee.
\newblock Improved baselines with visual instruction tuning, 2023{\natexlab{a}}.

\bibitem[Liu et~al.(2023{\natexlab{b}})Liu, Li, Wu, and Lee]{liu2023llava}
Haotian Liu, Chunyuan Li, Qingyang Wu, and Yong~Jae Lee.
\newblock Visual instruction tuning, 2023{\natexlab{b}}.

\bibitem[Liu et~al.(2024)Liu, Li, Li, Li, Zhang, Shen, and Lee]{liu2024llavanext}
Haotian Liu, Chunyuan Li, Yuheng Li, Bo Li, Yuanhan Zhang, Sheng Shen, and Yong~Jae Lee.
\newblock Llava-next: Improved reasoning, ocr, and world knowledge, 2024.

\bibitem[Loshchilov and Hutter(2017{\natexlab{a}})]{adamw}
Ilya Loshchilov and Frank Hutter.
\newblock Decoupled weight decay regularization.
\newblock \emph{arXiv preprint arXiv:1711.05101}, 2017{\natexlab{a}}.

\bibitem[Loshchilov and Hutter(2017{\natexlab{b}})]{cosine}
Ilya Loshchilov and Frank Hutter.
\newblock {SGDR}: Stochastic gradient descent with warm restarts.
\newblock In \emph{International Conference on Learning Representations}, 2017{\natexlab{b}}.

\bibitem[Masry et~al.(2022)Masry, Do, Tan, Joty, and Hoque]{masry-etal-2022-chartqa}
Ahmed Masry, Xuan~Long Do, Jia~Qing Tan, Shafiq Joty, and Enamul Hoque.
\newblock {C}hart{QA}: A benchmark for question answering about charts with visual and logical reasoning.
\newblock In \emph{Findings of the Association for Computational Linguistics: ACL 2022}, pages 2263--2279, Dublin, Ireland, 2022. Association for Computational Linguistics.

\bibitem[Mathew et~al.(2021)Mathew, Karatzas, and Jawahar]{docvqa}
Minesh Mathew, Dimosthenis Karatzas, and C.~V. Jawahar.
\newblock Docvqa: A dataset for vqa on document images.
\newblock In \emph{2021 IEEE Winter Conference on Applications of Computer Vision (WACV)}, pages 2199--2208, 2021.

\bibitem[Mathew et~al.(2022)Mathew, Bagal, Tito, Karatzas, Valveny, and Jawahar]{infovqa}
Minesh Mathew, Viraj Bagal, Rubèn Tito, Dimosthenis Karatzas, Ernest Valveny, and C.~V. Jawahar.
\newblock Infographicvqa.
\newblock In \emph{2022 IEEE/CVF Winter Conference on Applications of Computer Vision (WACV)}, pages 2582--2591, 2022.

\bibitem[Mishra et~al.(2019)Mishra, Shekhar, Singh, and Chakraborty]{mishraICDAR19}
Anand Mishra, Shashank Shekhar, Ajeet~Kumar Singh, and Anirban Chakraborty.
\newblock Ocr-vqa: Visual question answering by reading text in images.
\newblock In \emph{ICDAR}, 2019.

\bibitem[Nuriel et~al.(2022)Nuriel, Fogel, and Litman]{nuriel2022textadain}
Oren Nuriel, Sharon Fogel, and Ron Litman.
\newblock Textadain: Paying attention to shortcut learning in text recognizers.
\newblock In \emph{European Conference on Computer Vision}, pages 427--445. Springer, 2022.

\bibitem[Panagopoulou et~al.(2023)Panagopoulou, Xue, Yu, Li, Li, Joty, Xu, Savarese, Xiong, and Niebles]{panagopoulou2023xinstructblip}
Artemis Panagopoulou, Le Xue, Ning Yu, Junnan Li, Dongxu Li, Shafiq Joty, Ran Xu, Silvio Savarese, Caiming Xiong, and Juan~Carlos Niebles.
\newblock X-instructblip: A framework for aligning x-modal instruction-aware representations to llms and emergent cross-modal reasoning, 2023.

\bibitem[Raffel et~al.(2020)Raffel, Shazeer, Roberts, Lee, Narang, Matena, Zhou, Li, and Liu]{raffel2020exploring}
Colin Raffel, Noam Shazeer, Adam Roberts, Katherine Lee, Sharan Narang, Michael Matena, Yanqi Zhou, Wei Li, and Peter~J Liu.
\newblock Exploring the limits of transfer learning with a unified text-to-text transformer.
\newblock \emph{The Journal of Machine Learning Research}, 21\penalty0 (1):\penalty0 5485--5551, 2020.

\bibitem[Sidorov et~al.(2020)Sidorov, Hu, Rohrbach, and Singh]{textcaps}
Oleksii Sidorov, Ronghang Hu, Marcus Rohrbach, and Amanpreet Singh.
\newblock Textcaps: A dataset for image captioning with reading comprehension.
\newblock In \emph{Computer Vision -- ECCV 2020}, pages 742--758, Cham, 2020. Springer International Publishing.

\bibitem[Singh et~al.(2019)Singh, Natarajan, Shah, Jiang, Chen, Batra, Parikh, and Rohrbach]{textvqa}
Amanpreet Singh, Vivek Natarajan, Meet Shah, Yu Jiang, Xinlei Chen, Dhruv Batra, Devi Parikh, and Marcus Rohrbach.
\newblock Towards vqa models that can read.
\newblock In \emph{2019 IEEE/CVF Conference on Computer Vision and Pattern Recognition (CVPR)}, pages 8309--8318, 2019.

\bibitem[Tay et~al.(2022)Tay, Dehghani, Tran, Garcia, Bahri, Schuster, Zheng, Houlsby, and Metzler]{tay2022unifying}
Yi Tay, Mostafa Dehghani, Vinh~Q Tran, Xavier Garcia, Dara Bahri, Tal Schuster, Huaixiu~Steven Zheng, Neil Houlsby, and Donald Metzler.
\newblock Unifying language learning paradigms.
\newblock \emph{arXiv preprint arXiv:2205.05131}, 2022.

\bibitem[Wang et~al.(2022)Wang, Yang, Hu, Li, Lin, Gan, Liu, Liu, and Wang]{wang2022git}
Jianfeng Wang, Zhengyuan Yang, Xiaowei Hu, Linjie Li, Kevin Lin, Zhe Gan, Zicheng Liu, Ce Liu, and Lijuan Wang.
\newblock Git: A generative image-to-text transformer for vision and language, 2022.

\bibitem[Wang et~al.(2023)Wang, Wang, Lin, Bai, Zhou, Zhou, Wang, and Zhou]{wang2023onepeace}
Peng Wang, Shijie Wang, Junyang Lin, Shuai Bai, Xiaohuan Zhou, Jingren Zhou, Xinggang Wang, and Chang Zhou.
\newblock One-peace: Exploring one general representation model toward unlimited modalities, 2023.

\bibitem[Xue et~al.(2021)Xue, Constant, Roberts, Kale, Al-Rfou, Siddhant, Barua, and Raffel]{xue2021mt5}
Linting Xue, Noah Constant, Adam Roberts, Mihir Kale, Rami Al-Rfou, Aditya Siddhant, Aditya Barua, and Colin Raffel.
\newblock mt5: A massively multilingual pre-trained text-to-text transformer, 2021.

\bibitem[Yang et~al.(2021)Yang, Lu, Wang, Yin, Florencio, Wang, Zhang, Zhang, and Luo]{yang2021tap}
Zhengyuan Yang, Yijuan Lu, Jianfeng Wang, Xi Yin, Dinei Florencio, Lijuan Wang, Cha Zhang, Lei Zhang, and Jiebo Luo.
\newblock Tap: Text-aware pre-training for text-vqa and text-caption.
\newblock In \emph{Proceedings of the IEEE/CVF conference on computer vision and pattern recognition}, pages 8751--8761, 2021.

\bibitem[Zhang et~al.(2023{\natexlab{a}})Zhang, Li, and Bing]{zhang2023videollama}
Hang Zhang, Xin Li, and Lidong Bing.
\newblock Video-llama: An instruction-tuned audio-visual language model for video understanding, 2023{\natexlab{a}}.

\bibitem[Zhang et~al.(2023{\natexlab{b}})Zhang, Zhang, Gu, Zhou, Lipka, Yang, and Sun]{zhang2023llavar}
Yanzhe Zhang, Ruiyi Zhang, Jiuxiang Gu, Yufan Zhou, Nedim Lipka, Diyi Yang, and Tong Sun.
\newblock Llavar: Enhanced visual instruction tuning for text-rich image understanding, 2023{\natexlab{b}}.

\end{thebibliography}
}

\clearpage
\section*{Supplementary Materials for TAP-VL: Text Layout Aware Pretraining for Enriched Vision-Language Models}

\appendix
\section{Implementation Details}
\label{supp:implementation_details}
For all considered architectures, we used a uniform training procedure that involves applying LoRa \cite{hu2021lora} to the LLM and fine-tuning the vision projection module while keeping the ViT frozen. When implementing \methodname{}, we additionally pretrained and fine-tuned the OCR module during the \ptname{} and OCR-to-Language Alignment stages. Both the baseline models and \methodname{} were fine-tuned using the same mixture of datasets described in the next section (OCR-Vision-to-Language Alignment). All experiments were conducted on 8 Nvidia A100 (40G) GPUs using bfloat16 precision. The OCR system used in this paper is Textract OCR\footnote{https://aws.amazon.com/textract/}~\cite{litman2020scatter,aberdam2021sequence,nuriel2022textadain,aberdam2022multimodal,aberdam2023clipter}.

\subsection{Training Datasets}
\label{ssup:training datasets}

For our experiments, we employed two distinct dataset combinations for the OCR-to-Language Alignment and OCR-Vision-to-Language Alignment phases. The datasets used in these phases, along with their evaluation metrics and splits, are detailed in \cref{tab:datasets}.

\noindent \textit{\textbf{OCR-to-Language Alignment:}} This phase was dedicated to OCR-centric VQA datasets, where answers can be directly inferred from the OCR text including: DocVQA, InfoVQA, ChartQA, OCRVQA, TextVQA, and STVQA.

\noindent \textit{\textbf{OCR-Vision-to-Language Alignment}} was trained on a combination of OCR-centric and non-OCR-centric datasets, specifically DocVQA, InfoVQA, ChartQA, OCRVQA, TextVQA, STVQA, TextCaps, COCO, and VQAv2.

\subsection{Training hyperparameters}
In each stage of \methodname{}'s training, the OCR module consistently generated 32 query tokens. Additionally, the AdamW optimizer \cite{adamw} and the Cosine Annealing scheduler \cite{cosine} were uniformly applied. Beyond these constants, each stage was characterized by its own distinct set of hyperparameters

\begin{table}[ht!]
\centering
\resizebox{0.5\textwidth}{!}{
\begin{tabular}{l}
\hline
\textbf{Template} \\
\toprule
Could you write a short image caption? \\
Could you write a short image description?\\
What does this image show? \\
Could you write a short description for the image? \\
Could you write a description for the photo? \\
Could you provide a description of what is presented in the photo? \\
Could you briefly describe the content of the image? \\
Can you briefly explain what you see in the image? \\
Could you use a few words to describe what you perceive in the photo? \\
Could you provide a short depiction of the picture? \\
Could you use a few words to illustrate what is happening in the picture? \\
\bottomrule
\end{tabular}}
\caption{\textbf{Instruction Templates for Captioning.} Overview of templates employed across captioning datasets to guide caption generation.}

\label{tab:caption_templates}
\end{table}

\begin{table*}[htb]
  \centering
  \small 
  \setlength{\tabcolsep}{5pt} 
    \resizebox{\textwidth}{!}{%

  \begin{tabular}{l|c|c|c|c|c}
    \toprule
    Task & Dataset & Description & Train split &  Eval split &  Metric\\
    \hline
    \multirow{3}{*}{Scene-Text VQA} & TextVQA \cite{textvqa}& Text-oriented VQA on natural images& train& val & vqa-score(↑)\\
    & STVQA \cite{stvqa}& Text-oriented VQA on natural images& train&test & ANLS(↑)\\
    & OCRVQA  \cite{mishraICDAR19}& Text-oriented VQA on natural images&train& - & Acc@1(↑)\\
    \hline
    \multirow{4}{*}{Document Understanding} & DocVQA \cite{docvqa}& VQA on single page scanned documents&train& test & ANLS(↑)\\
    & InfoVQA \cite{infovqa}& VQA on infographic images& train& test & ANLS(↑)\\
    & ChartQA \cite{masry-etal-2022-chartqa}& VQA on chart images&train (human)& - & RA\\
    & Dude \cite{dude}& VQA on multipage scanned documents& -&test & ANLS(↑)\\
    \hline
    Image Caption & COCO \cite{chen2015microsoft}& Captioning of natural images &train& test & CIDEr(↑) \\
    \hline
    Scene-Text Caption & TextCaps \cite{textcaps} & Text-oriented Captioning of natural images &train& test & CIDEr(↑) \\
    \hline
    General VQA & VQAv2 \cite{vqav2}& VQA on natural images &train& val & vqa-score(↑)\\
    \bottomrule
  \end{tabular}
  }

    \caption{\textbf{Datasets} used during the finetuning stages.}

    \label{tab:datasets}
\end{table*}

\noindent \textit{\textbf{\ptname:}} The pretraining stage comprised 140,000 training steps, with a learning rate \(1e-4\) and and a batch size of $224$. The OCR module was trained with a maximum of 512 token in the OCR module and a masking probability of 0.15. More information about the pretraining optimization can be found in \cref{tab:implementation1}.
\setlength{\tabcolsep}{4pt}
\begin{table*}[h!]
    \begin{center}
        
        \resizebox{\textwidth}{!}{
        \begin{tabular}{lcccccccccc}
            \toprule
             Stage &\# Steps & Batch Size & Base LR & \# Warmup steps & Weight Decay & \qfname{} Prompt & \# OCR module token& Mask density\\
            \midrule
           
             Pretraining&$140K$ & $224$ & $1e-4$ &  $1000$ & $0.05$ & \texttt{<extra\_id\_i>} \{\texttt{masked\_words\_i}\}&512& 0.15\\  
             Alignment&$300K$&$24\textsuperscript{$\dagger$}$&$2e-5$&$1000$&$0.05$&{\texttt{Question}: \{\texttt{Instruction}\}}&1024&0\\
            \bottomrule
        \end{tabular}
        }
    \caption{\textbf{Model hyperparameters used during the \ptname{} and OCR-to-language Alignment.} \textsuperscript{$\dagger$} Qwen-VL and LLaVA was trained using a batch size of 32 during the OCR-to-language Alignment. }
        \label{tab:implementation1}

    \end{center}
    
    \end{table*}
\setlength{\tabcolsep}{1.4pt}

\noindent \textit{\textbf{OCR-to-Language Alignment:}} Detailed in \cref{tab:implementation1}, this stage maintained a uniform structure across models, with $300K$ training steps, $1000$ warmup steps, a learning rate of $2e-5$, and a batch size of 24 for InstructBlip and 32 for the other models. The OCR branch's maximum token length was set to 1,024 for this training phase.

\noindent \textit{\textbf{OCR-Vision-to-Language Alignment:}} In \cref{tab:implementation_table_stage3}, we present the hyperparameters for the OCR-vision-to-language alignment phase. This phase mirrored the prior alignment stage's training steps, batch size, warmup procedure, and learning rates. During the training phase, the OCR module was trainable for all models except Qwen-VL. The image resolution used to feed the frozen vision encoder was the original one used by the baseline models, with InstructBlip XL and XXL set at 224, LLaVA-1.6 at 336, and Qwen-VL at 448. The maximum OCR module length was set to 2,000 tokens for \methodname{}\textsubscript{InstructBlip XL} and 1,400 for the other configurations. The LLM prompt length was constrained by RAM limitations, set to different maximums tailored to each model configuration.

\setlength{\tabcolsep}{4pt}
\begin{table*}[h!]
    \centering
    \small
    \setlength{\tabcolsep}{5pt}
    
    \resizebox{\textwidth}{!}{
        \begin{tabular}{lcccccccc}
            \toprule
            VL Model & \# Steps &\# LLM token& \# OCR module token &  LoRA ($\alpha,r,dropout, modules$) & OCR module & Image resolution\\
            \midrule
            InstructBlip&$300K$&$1024$&$2000$&$16,32,0.05, [W_q,W_v]$& Trained & 224\\
            InstructBlip XXL&$300K$&$400$&$1024$&$16,32,0.05, [W_q,W_v]$&Trained& 224\\
            LLaVA-1.6&$300K$&$400$&$1024$&$16,32,0.05, [W_q,W_v]$&Trained&336\\
            Qwen-VL&$100K$&$600$&$1400$&$16,32,0.05, [W_q,W_v]$&Frozen&448\\
            \bottomrule
        \end{tabular}
    }
    \caption{\textbf{Hyperparameters for OCR-Vision-to-Language Alignment.} Parameters such as batch size, learning rate, warm-up period, and optimizer are not specified here, as they remain consistent with those used in the OCR-to-Language Alignment stage.}
        \label{tab:implementation_table_stage3}

\end{table*}
\setlength{\tabcolsep}{1.4pt}

\subsection{Instruction templates}
For the VQA-based datasets, we use the given question as the instruction. For the captioning datasets, we randomly select an instruction that asks the model to describe the image among the one in \cref{tab:caption_templates}.

\subsection{Pretraining data preparation:}
Each document contains OCR tokens, denoted as $\evt_1, \evt_2, \ldots, \evt_n$ with corresponding bounding boxes $\rvb_1, \rvb_2, \ldots, \rvb_n$. To create training pairs, we randomly mask spans of OCR tokens along with their positional information. Specifically, we: 
 
 \begin{itemize}
     \item Sample $M$ spans, each defined by a start index $s_i$ and an end index $e_i$.
     \item Replace tokens and bounding boxes in each span $(s_i, s_i+1, \ldots, e_i)$ with a special token \texttt{<extra\_id\_i>} and the minimal bounding box covering the span, respectively (following the method in \cite{latr}).
     \item Generate pairs consisting of the noisy OCR input and the masked words, i.e., the original tokens in the masked spans. The masked words are represented as a sequence $ (\texttt{<extra\_id\_i>}$ $\texttt{value\_i>})_{i=1}^{i=M}$ where \texttt{value\_i} is the original text in the $i$-th masked span.
      \end{itemize}

\subsection{Layout-Aware pretraining}
\label{ssup:pretraining}
For all the pretraining objectives, the OCR encoder produces rich embeddings from the noisy OCR. We denote these embeddings as $O \in \R^{B \times l \times d_{\text{ocr}}}$, where $B$ is the batch size, $l$ is the number of noisy OCR tokens, and $d_{\text{ocr}}$ is the dimensionality of the embeddings. Additionally, we define $\rmR_q \in \R^{B \times K \times d}$ to represent the learnable queries, where $B$ is the batch size, $K$ is the number of learnable queries, and $d$ is their dimensionality. The representation of the $M$ mask words, preceded by the task specific special token, is denoted as $\rmR_m \in \R^{B \times (2M+1) \times d}$.

\textbf{\ptone{}}: 
In this pretraining objective, we use a \texttt{<dec>} special token.
The \qfname{} processes $\rmR_q$ through its self-attention ($SA$) layers, while processing  $\rmR_m$ using causal self-attention (CSA) layers conditioned on $\rmR_q$. This results in mask-words representations that integrate contextual information from the queries tokens. Subsequently, the queries representations are updated using the noisy OCR content using a cross-attention (CA) layer. Two finals feed-forward (FF) layers are applied on top of $\rmR_q$ and $\rmR_m$.
Formally, we compute: 
\begin{equation}
    \rmR_q^{(out)} = \text{FF}\left(\text{CA}\left(\text{SA}\left(\rmR_q^{(in)} \right), O \right) \right)
    \end{equation}
\begin{equation}
\rmR_m^{(out)} = \text{FF}\left(\text{CSA}\left(\rmR_m^{(in)} \mid \rmR_q^{(in)}\right)\right)
\end{equation}

where $(in)$ and $(out)$ represent the input and output representations.

We then apply a language modeling loss $\mathcal{L}_{\text{LM}}$ over the output representations $\rmR_m^{\text{(out)}}$ to recover the original masked content. Specifically, we pass $\rmR_m^{\text{(out)}}$ through a softmax function to obtain the predicted token probabilities: \begin{equation} \hat{y}_i = \text{Softmax}\left(\rmR_{m}^{i (\text{out})}\right) \end{equation}

The language modeling loss $\mathcal{L}_{\text{LM}}$ is then computed using the cross-entropy between the predicted probabilities and the ground truth tokens $(y)_{i=1}^M$: 

\begin{equation} \mathcal{L}_{\text{LM}} = - \frac{1}{B} \sum_{i=1}^{B} \sum_{j=1}^{M} \log \left( \hat{y}_{j}|y_{t<j} \right). \end{equation}

\textbf{\pttwo{}:} 
In this objective, we use a
the \texttt{<cls>} token and its specific representation is denoted as $\rmR_t \in \R^{B \times 1 \times d}$.

The \qfname{} processes $\rmR_q$ and $\rmR_m$ using two independent self-attention layers. This results in mask-words representations that are independent from the from the queries tokens representation. Subsequently, the queries representations are updated using the noisy OCR content using a cross-attention layer. Two finals feed-forward layers are applied on top of $\rmR_q$ and $\rmR_m$.
Formally, we compute: 

\begin{equation}
    \rmR_q^{(out)} = \text{FF}\left(\text{CA}\left(\text{SA}\left(\rmR_q^{(in)} \right), O \right) \right)
    \end{equation}
\begin{equation}
\rmR_m^{(out)} = \text{FF}\left(\text{SA}\left(\rmR_m^{(in)}\right)\right)
\end{equation}
The $\rmR_t^{(out)}$ representation is then extracted from $\rmR_m^{(out)}$ in order to compute
the pairwise similarity between $\rmR_q^{(out)}$ and $\rmR_t^{(out)}$, yielding $\rmP_{qt} \in \R^{B \times B \times K}$, and subsequently, the maximum similarity is selected across the last dimension, resulting in $\rmS_{qt} 
\in \R^{B \times B}$. 
Finally, we apply the contrastive learning loss~\cite{chen2020simple}, with a temperature scalar $\tau$ on the  $\rmS_{qt}$ matrix:
\begin{equation}
    \mathcal{L}_{\text{Contrastive}}(\rmS) = -\frac{1}{B} \sum_{i=1}^B \log \left( \frac{\exp(\rmS_{ii}/\tau)}{\sum_{j \ne i} \exp(\rmS_{ij}/\tau)} \right)
\end{equation}

\textbf{\ptthree{}:}
In this objective, we compute $\rmR_q^{\text{(out)}}$, $\rmR_m^{\text{(out)}}$, and $\rmS_{qt}$ similarly to \ptthree{}. We employ a hard negative mining strategy \cite{li2021align} to select challenging negative examples based on the similarity values in the $\rmS_{qt}$ matrix. This approach yields pairs of representations $\rmR_{q}^{i\text{(out)}}$ and $\rmR_{m}^{j\text{(out)}}$ where $i \ne j$, indicating they come from different documents. Additionally, we consider pairs where $i = j$, which represent positive examples originating from the same document.
 The query representation $\rmR_q^{\text{(out)}}$ is projected to $\rmR_q \in \rmR^{B \times 1 \times 1}$ using a feed-forward layer followed by average pooling across the query dimension. This provides a single similarity value for each pair of noised OCR-masked words. Finally, we apply a binary cross-entropy loss to encourage the model to correctly detect matching pairs, where $y_i = 1$ if the pair matches and $0$ otherwise.
\begin{equation}
    \mathcal{L}_{\text{BCE}} = -\frac{1}{B} \sum_{i=1}^B y_i \log \left( \sigma\left(\rmP_q^i\right) \right) + (1 - y_i) \log \left( 1 - \sigma\left(\rmP_q^i\right) \right)
\end{equation}

\section{Additional results}
\label{ssup:additionalresults}

\begin{figure*}[htb!]
  
  \includegraphics[width=1\linewidth]{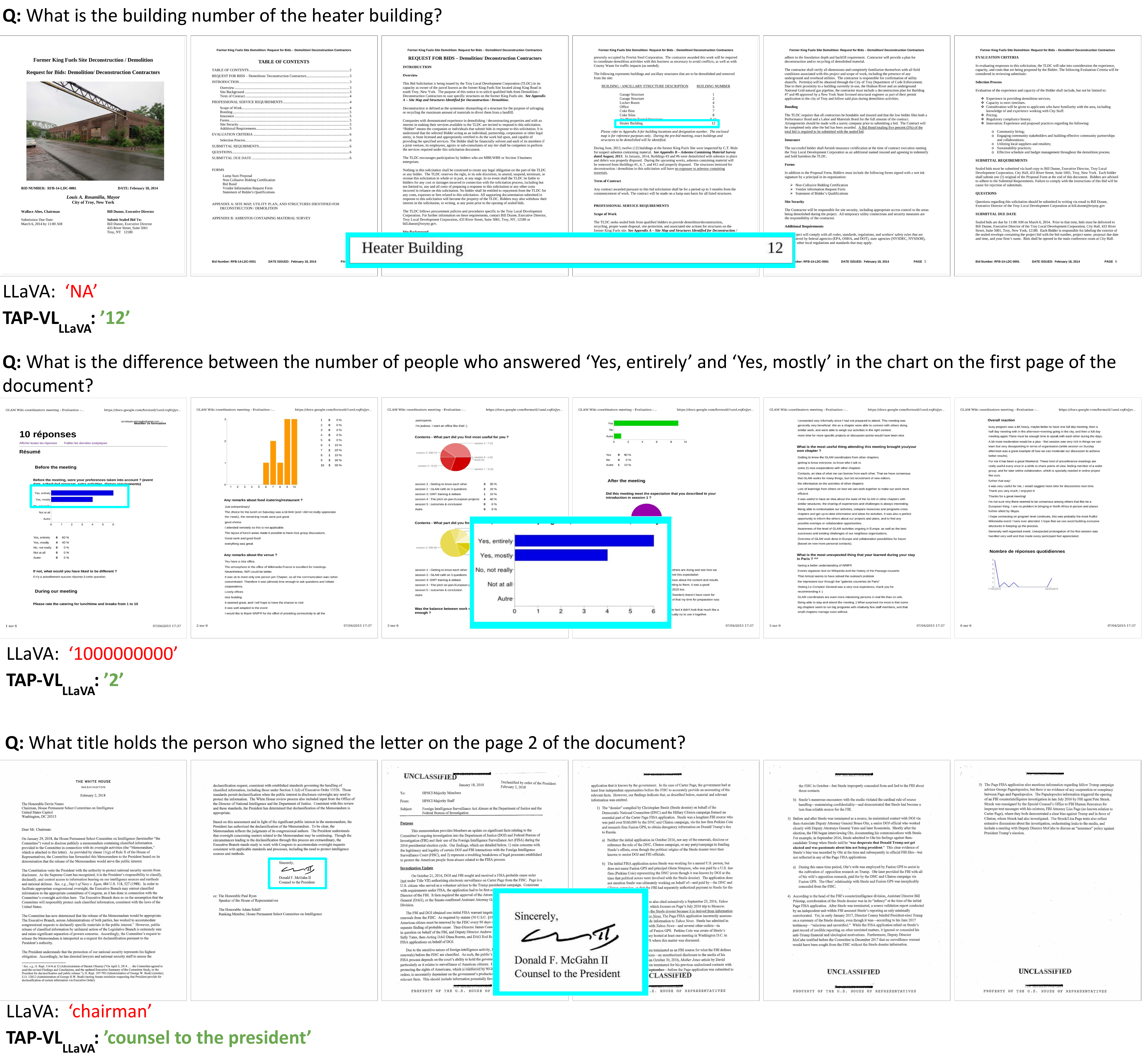}
  \caption{\textbf{Qualitative Improvements Demonstrated by \methodname{}}. Illustrative examples showcasing the improvements achieved by our method on multipage document VQA benchmarks using LLaVA. \methodname{} enhances the base model's ability to grasp OCR and layout information, yielding significant improvements across both benchmark types.}
    \label{fig:qualitative2}

\end{figure*}

\begin{figure*}[htb!]
  
  \includegraphics[width=1\linewidth]{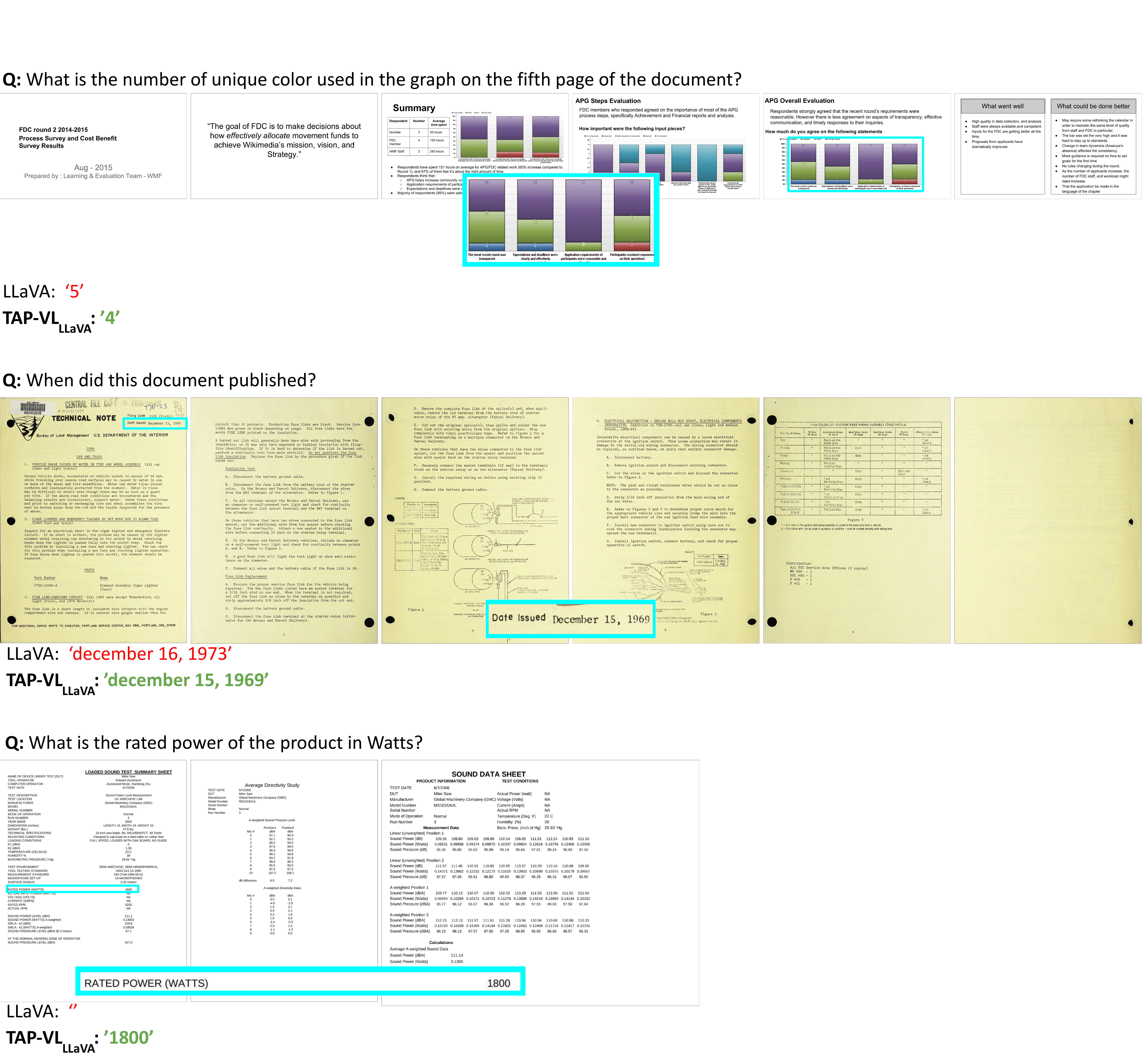}
  \caption{\textbf{Qualitative Improvements Demonstrated by \methodname{}}. Illustrative examples showcasing the improvements achieved by our method on multipage document VQA benchmarks using LLaVA. \methodname{} enhances the base model's ability to grasp OCR and layout information, yielding significant improvements across both benchmark types.}
    \label{fig:qualitative3}

\end{figure*}

\noindent \textit{\textbf{Multipage qualitative results}}
In \cref{fig:qualitative2,fig:qualitative3}, we display how our method integrates into LLaVA-1.6 to leverage the information available inside a multiple page document in order to answer a given question. For instance, the base model struggles to identify the "\textit{number of the heater building}" located in the fourth page (over six), whereas \methodname{} effectively uses layout information to understand it.

\end{document}